\documentclass[lettersize,journal]{IEEEtran}
\usepackage{amsmath,amsfonts}
\usepackage{algorithmic}
\usepackage{algorithm}
\usepackage{array}
\usepackage[caption=false,font=normalsize,labelfont=sf,textfont=sf]{subfig}
\usepackage{textcomp}
\usepackage{stfloats}
\usepackage{url}
\usepackage{verbatim}
\usepackage{graphicx}
\usepackage{cite}
\usepackage{multirow}
\usepackage{makecell}
\usepackage{setspace}
\usepackage{amsmath, epsfig}
\usepackage{subfig}
\usepackage{subfloat}
\begin{document}

\title{Unsupervised Domain Adaptive Learning via Synthetic Data for Person Re-identification}

\author{Qi~Wang,~\IEEEmembership{Senior Member,~IEEE,}
	Sikai~Bai,~
	Junyu~Gao,~\IEEEmembership{Member,~IEEE,}
	and~Yuan~Yuan,~\IEEEmembership{Senior Member,~IEEE,}
\thanks{Q. Wang, S. Bai, J. Gao, Y. Yuan and X. Li are with the School of Artificial Intelligence, OPtics and Electronics (iOPEN), Northwestern Polytechnical University, Xi'an 710072, P.R. China. E-mail: crabwq@gmail.com, whitesk1973@gmail.com, gjy3035@gmail.com, y.yuan1.ieee@gmail.com.}
}

\markboth{IEEE Transactions on Image Processing}%
{Shell \MakeLowercase{\textit{et al.}}: A Sample Article Using IEEEtran.cls for IEEE Journals}


\maketitle

\begin{abstract}
Person re-identification (re-ID) has gained more and more attention due to its widespread applications in intelligent video surveillance. Unfortunately, the mainstream deep learning methods still need a large quantity of labeled data to train models, and annotating data is an expensive work in real-world scenarios. In addition, due to domain gaps between different datasets, the performance is dramatically decreased when re-ID models pre-trained on label-rich datasets (source domain) are directly applied to other unlabeled datasets (target domain). In this paper, we attempt to remedy these problems from two aspects, namely data and methodology. Firstly, we develop a data collector to automatically generate synthetic re-ID samples in a computer game, and construct a data labeler to simultaneously annotate them, which free humans from heavy data collections and annotations. Based on them, we build two synthetic person re-ID datasets with different scales, namely, “GSPR” and "mini-GSPR" datasets. Secondly, we propose a synthesis-based multi-domain collaborative refinement (SMCR) network, which contains a synthetic pretraining module and two collaborative-refinement modules to implement sufficient learning for the valuable knowledge from multiple domains. Extensive experiments show that our proposed framework obtains significant performance improvements over the state-of-the-art methods on multiple unsupervised domain adaptation tasks of person re-ID.
\end{abstract}


\section{Introduction}
\IEEEPARstart{P}{erson} re-identification(Re-ID) aims at identifying images of the same pedestrian across non-overlapping camera views in different places, which has attracted a lot of research interests since the urgent demand for public safety and the increasing number of surveillance cameras. Benefiting from the development of deep learning \cite{he2016deep, szegedy2015going} and the availability of labeled re-ID datasets \cite{zheng2015scalable, ristani2016performance, wei2018person}, CNN-based re-ID methods \cite{ye2020deep, lin2020unsupervised, jiang2020self, yao2019deep}, have made remarkable performance improvements in a supervised manner. However, the aforementioned approaches need a multitude of accurately labeled and diversified data to learn the discriminative features, and the current datasets are not able to perfectly satisfy the demands in dataset scale or data diversity. 
\begin{table}[h]
	\setlength{\abovecaptionskip}{0.cm}
	\begin{spacing}{1.1}
		\begin{center}
			\caption{ Statistic of real-world and synthetic person re-ID datasets. "View" denotes whether the dataset has viewpoint labels }\label{table4}
			\resizebox{1.0\hsize}{!}{
				\begin{tabular}{c| c| c| c| c| c}
					\hline
					\multicolumn{2}{c|}{Datasets} &{\#box} &{\#identity} &{\#cam} &{view} \\
					\hline
					\multirow{4}{*}{\rotatebox{90}{real data} } 
					& Market-1501 &32,688 &1,501 &6 &N \\
					& Market-1203 &8,569 &1,203 &2 &N \\
					& CUHK03 &14,096 &1,467 &2 &N\\
					& DukeMTMC-ReID &36,441 &1,812 &8 &N\\ 
					& MSMT17 &126,441 &4,101 &15 &N \\ \hline
					
					\multirow{6}{*}{\rotatebox{90}{synthetic data} } 
					& SOMAset &100,000 &50 &- &N \\
					& SyRI &1,680,000 &100 &- &N \\
					& PersonX &273,456 &1,26 &6 &Y\\
					& GPR &443,352 &754 &12 &Y\\ 
					& \textbf{mini-GSPR} &27,040 &520 &13 &Y \\ 
					& \textbf{GSPR} &\textbf{625,416} &\textbf{2,369} &\textbf{26} &Y \\ \hline
					
				\end{tabular}\label{datasets}
			}
		\end{center}
	\end{spacing}
\end{table}
For example, Market1501 \cite{zheng2015scalable} is only collected by six cameras during the summer vacation, and DukeMTMC-reID \cite{ristani2016performance} only contains samples captured by eight cameras in frigid scenes. Meanwhile, all data contained in these datasets are only recorded on the university campus. It causes a tricky problem that the re-ID models trained on these datasets produce a large performance degradation in unseen extreme cases (e.g., different viewpoints, variant low-image resolutions, illumination changes, \emph{etc}). Moreover, because it may violate people's privacy, the collection and annotation of re-ID data collection and annotation also become more and more difficult in real world. Regardless of the infringement of people's privacy, it is impractical to continuously annotate person identities in various new target domains.

In order to address the aforementioned issues, many attempts have been made in terms of data and method. Specifically, from the perspective of data, some approaches have been proposed to make use of commercial software to construct synthetic re-ID datasets \cite{sun2019dissecting, barbosa2018looking, xiang2020attribute}. Leveraging synthetic datasets is a meaningful attempt to remedy the reliance on large-scale real datasets, and the constructed datasets are usually utilized to provide the effective initialization for re-ID models. However, as shown in Table \ref{datasets}, existing synthetic data sets suffer from limited data volume or scarce diversity to some extent, and there exist serious differences between synthetic and real datasets due to unrealistic personal characteristics and scenes. It leads to sub-optimal initialization, and there is a poor performance when directly performing the domain adaptive person re-ID tasks from the synthetic datasets to the real-world scenarios.     

From the perspective of methodology, unsupervised domain adaptation (UDA) for person re-ID has been proposed, which aims at tackling domain discrepancy between different datasets. There are two main research directions for this problem. The first are domain translation-based methods, which focus on the translation from source domain to target domain by resorting to generating source-to-target samples \cite{wei2018person, deng2018similarity, chen2019instance}, where the generated samples inherit ground-truth identity labels from the source domain data and have a similar style with the target domain data.

However, source domain knowledge is simply abandoned after domain translation, and there are no auxiliary pseudo-labels for target domain data in these approaches. The second ones are pseudo label-based methods, which alternately generate pseudo labels by clustering samples from the target domain and train the network with generated pseudo-labeled data \cite{yu2019unsupervised, zhang2019self, huang2020proxy, zhai2020ad}. Although this pipeline has achieved remarkable performance, noisy pseudo-labels and large domain gaps between source and target domains hinder performance further improvements.

In this paper, we attempt to remedy the above problems from two aspects, namely data and methodology. From the data view, we develop a data collector and labeler with the help of a popular electronic game "Grand Theft Auto V" (abbreviation as GTA V), which can be used to generate synthetic pedestrian samples and automatically annotate samples. Based on them, we first build a large-scale synthetic person re-ID dataset, named as “GTA V Synthetic Person Re-ID” (GSPR) dataset. Compared with existing person re-ID datasets, our GSPR has the following advantages: 1) This is a more large-scale person re-ID dataset that consists of 625416 images and a total of 2369 identities across 26 non-overlapping cameras. 2) It is free of collection and annotation, which can liberate people from the labor-intensive collection and annotation of large amounts of data. 3) It is a diversified dataset and considers various extreme cases in real-world scenarios, such as multiple viewpoints, varying image resolutions, illumination changes, unconstrained poses, occlusions, \emph{etc}. Moreover, we construct a mini-scale person re-ID dataset, called  “mini GTA V Synthetic Person Re-ID” (mini-GSPR) dataset, which has high resolution and only considers various ideal visual factors. The detailed statistics are illustrated in Table 1.

From the methodological view, we propose a synthesis-based multi-domain collaborative refinement (SMCR) network to improve the performance in domain adaptive person re-ID tasks. There are three strategies in the SMCR network to achieve the goal. Firstly, conventional re-ID methods are pre-trained on ImageNet or synthetic datasets. Due to severe differences between synthetic and real datasets and monotonous pretraining in a single domain, they might be not the optimum option for domain adaptive tasks. To this end, we design a novel pretrained scheme, which exploits the constructed synthetic datasets to pre-train the domain adaptive model by combining source domain data. Considering the serious shift between synthetic and real datasets, we generate synthetic-to-source samples by performing domain translation between synthetic data and source domain. Essentially, our scheme regards the pretrained process as the performance generalization from synthetic-to-source samples to the source domain, and it is able to provide better initialization for domain adaptive models.

Furthermore, we construct two synchronous training and collaborative refinement modules, namely domain translation hybrid refinement (DTHR) module and relation invariant hybrid refinement (RIHR) module. The devised two modules aim to reduce the negative impacts of the domain gap and preserve original information in the source domain, respectively. DTHR contains a domain translator, a feature encoder, and a pseudo-label generator, which integrates the processes of domain translation and pseudo label prediction, while performing joint training with the generated style-transferrer samples and target domain data. RIHR contains a feature encoder and a pseudo-label generator, which aims to maintain the original inter-instance relations invariant and implement joint learning for the source and target data. Finally, due to the similar architecture and complementary roles between DTHR and RIHR, we introduce a collaborative refinement strategy to further optimize the generated pseudo-labels and the discrimination of target features. Besides, we are sorting out the proposed synthetic datasets and the source code for SMCR, which will be released later.

The major contribution of our work can be summarized as three-fold:

\begin{enumerate}
	\item[1)] \textbf{Datasets.} Two different scale synthetic person re-ID datasets (i.e., GSPR and mini-GSPR) are constructed with the help of our designed data collector and labeler. There is no labor-intensive cost in the data generation pipeline. Compared with the existing person re-ID datasets, GSPR is a more large-scale and diversified synthetic dataset, and mini-GSPR has higher resolution and more ideal visual conditions. Our datasets effectively relieve the dependence on large-scale real labeled datasets and alleviate the difficulty of domain adaptation.
	\item[2)] \textbf{Methodology.} We propose a synthesis-based multi-domain collaborative refinement (SMCR) network to sufficiently exploit the valuable knowledge from multiple domains, which includes a synthetic pretraining module and two collaborative-training modules. The SCMR network not only reduce the adverse impacts of domain discrepancy, but provides a novel learning scheme by utilizing synthetic data for domain adaptive tasks.
	\item[3)] \textbf{Experiments.} Extensive experiments are performed on multiple unsupervised domain adaptation tasks of person re-ID, which demonstrate the proposed SMCR obtains consistently optimal performance to the state-of-the-art approaches. We perform ablation studies to verify that all the proposed components contribute to domain adaptive performance. 
\end{enumerate}


\section{Related Works}
\label{sec:relate}
In this section, we first review the previous works about synthetic data, and then briefly describe unsupervised domain adaptation and its applications on person re-ID.
\subsection{Synthetic Data}
Manually collecting and labeling an ideal dataset in the real environment is extremely time-consuming and labor-intensive, and it is probably to violate people's privacy. Because unlimited samples can be generated by utilizing sythetic data collector and labeler in theory, many attempts have been made to construct synthetic datasets and alleviate the dependence on large-scale real labeled datasets. This strategy has been applied in problems like semantic segmentation \cite{mccormac2017scenenet, sankaranarayanan2018learning}, pose estimation \cite{yuan2019bio, yuan2019memory}, crowd counting \cite{wang2019learning, wang2020pixel}, \emph{etc}. In the person re-ID, \cite{barbosa2018looking} proposed a synthetic instance dataset SOMAset by resorting to photorealistic human body generation software, which contains 50 person models and 11 types of outfits. \cite{bak2018domain} introduced SyRI dataset using 100 virtual humans illuminated with rich HDR environment maps. In addition, \cite{sun2019dissecting} presented a large-scale synthetic dataset named PersonX, which has 1,266 hand-crafted identities and 6 different scenes with multiple viewpoints. Recently, \cite{xiang2020unsupervised} constructed a GPR dataset, which includes 754 indenties across 12 non-overlapping cameras. In GPR dataset, the number of cameras and the data volume are still not enough, and the author only designed a coarse data filter based on this dataset to manually select some specific scenes for different datasets. Moreover, the above synthetic datasets are either uneditable or not released to the public.

Comparatively, the proposed synthetic datasets (i.e., GSPR and mini-GSPR) have more diversified scenes, identity categories and data volume. Moreover, various visual factors contained in the constructed datasets can be freely edited/extend, including scenes, weathers, illuminations, data volume, \emph{etc}, which are not limited to this specific study, and can also be extended to future research in other areas. 

\subsection{Unsupervised Domain Adaptation}
Unsupervised domain adaptation (UDA) defines a learning problem, where the learned knowledge from the labeled source domain is generalized to the target domain with unknown classes, in order to properly measure the inter-instance affinities of the target domain. Early approaches focused on feature/sample mapping between source and target domains \cite{sun2015return, liu2016coupled}. The respective work, correlation alignment (CORAL) \cite{sun2015return}, aligned the covariance and mean between source- and target-domain distributions. Furthermore, some methods aim to find domain-invariant feature spaces \cite{ajakan2014domain, ganin2016domain}. \cite{long2015learning} and \cite{tzeng2014deep} designed the Maximum Mean Discrepancy (MMD) to extract domain-invariant features. \cite{ganin2016domain} and \cite{ajakan2014domain} proposed domain confusion loss for this purpose.

Recently, some researchers introduced adversarial learning and pseudo-label generation strategies to reduce the differences between data distributions of source and target domains. Adversarial learning based methods \cite{bousmalis2017unsupervised, hoffman2018cycada} introduced a cycle consistency loss by DiscoGAN \cite{kim2017learning}, DualGAN \cite{yi2017dualgan}, and CycleGAN \cite{zhu2017unpaired} to train a domain generator and discriminator to eliminate the discriminative domain information from the learned features. CYCADA \cite{hoffman2018cycada} transferred the input samples across domains at both pixel- and feature-level. The pseudo-label generation based approaches \cite{chen2011co, sener2016learning, zhang2018collaborative} focused on modeling relations between unlabeled target domain instances using the predicted pseudo-label. However, the above approaches only focused on conventional domain adaptation for close-set recognition, where the source and target domains share the same label set, and they are not suitable for unsupervised domain adaptation on person re-ID.

\subsection{UDA on person re-ID}
Unsupervised domain adaptation (UDA) for person re-ID is an open-set problem, which has disjoint person identity systems between source and target domains. There are extensive studies in unsupervised domain adaptive person re-ID. Domain alignment-based methods tried to minimize the data distribution of the source and the target domain. \cite{lin2018multi} utilized Maximum Mean Discrepancy (MMD) distance to align the distributions of the source’s and the target’s mid-level features. \cite{wang2018transferable} applied additional attribute annotations to align feature distributions in common space. Furthermore, domain translation-based methods applied generative adversarial network to generate style-transferred samples with a similar style to the target domain, where the re-ID model is pre-trained in labeled style-transferred samples to perform finetuning in the target domain. PTGAN \cite{wei2018person} and SPGAN \cite{deng2018image} maintained the ID-related features invariant by utilizing identity-based regularization. SDA \cite{ge2020structured} used online relationship consistency regularization term to maintain inter-sample relations. Moreover, ATNet \cite{liu2019adaptive}, CR-GAN \cite{chen2019instance} and PDA-Net \cite{li2019cross} are effective to transfer images with identity labels from source into target domains to learn domain-invariant features. These methods aimed to preserve the original IDs or inter-sample relations during translation, and provide a plausible way to reduce domain gaps between source and target domain by generating style-transferred samples. But the source domain data is simply discarded after translation, and there are no pseudo-labels for target domain data in these strategies. In fact, experimental evidence in \cite{ge2020structured} and \cite{ge2020self} has verfied that original inter-sample relations in the source domain and pseudo labels in the target domain are indispensable for cross-domain person re-ID tasks.
\begin{figure*}
	\begin{center}
		\centering
		\includegraphics[width=0.96\linewidth]{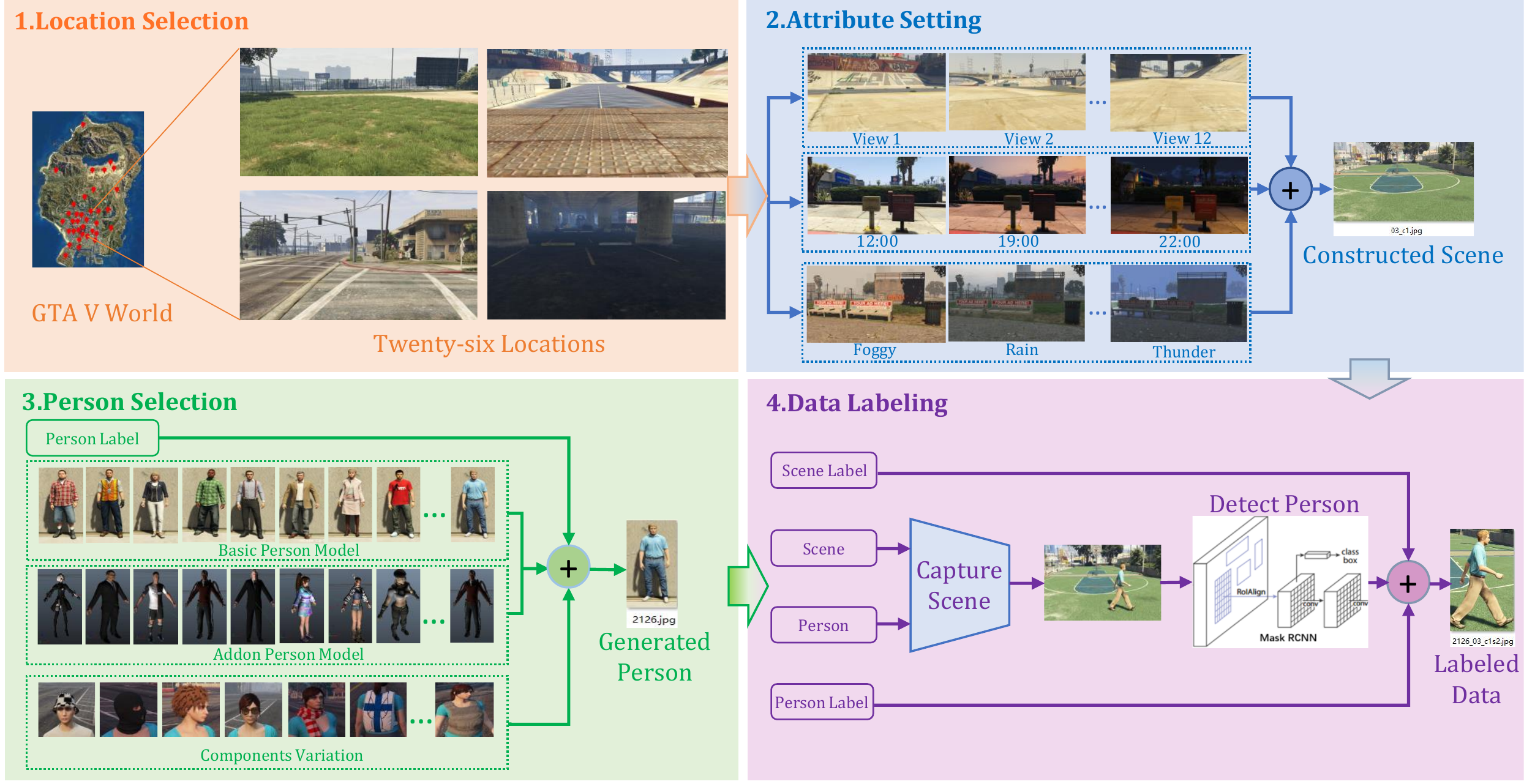}
		\hfill
	\end{center}
	\caption{The pipline of data generation in GTA 5, including location selection, attribute setting, person selection and data labeling.}\label{data_overview}
\end{figure*}

The pseudo-label-based approaches \cite{wang2020unsupervised, fu2019self, ge2020mutual, ge2020self} performed feature learning in the unlabeled target domain by utilizing the generated pseudo labels, which have attracted much
attention since the very limited cost of manual annotations and state-of-the-art performance. Specifically, the pseudo labels is able to be generated by either clustering features or measuring similarities with exemplar features. PUL \cite{fan2018unsupervised} designed a cluster and self-pace training framework. SSG \cite{fu2019self} assigned multi-scale pseudo labels by utilizing human local features. PAST \cite{zhang2019self} introduced a progressive training strategy. AD-Cluster \cite{zhai2020ad} incorporated style-translated sample into clustering to improve the discriminativeness of instance features. Moreover, MMT \cite{ge2020mutual} and SPCL \cite{ge2020self} introduced mutual learning and self-paced learning respectively to further optimize the pseudo labels. Nevertheless, the above methods cannot sufficiently exploit all valuable knowledge from multiple domains. Most of these methods suffer from the adverse impacts of domain gap and noisy pseudo label. 

In addition to considering the discrepancy between different domains, there are some works to camera differences within the single domain. \cite{wu2019unsupervised} presented a camera-aware similarity consistency loss to learn consistent pairwise similarity distributions for intra-camera matching and cross-camera matching. \cite{ristani2016performance} develop amera-aware domain adaptation framework to reduce the discrepancy between both domains and cameras. \cite{zhong2018camera} introduced camera style adaptation to produce images with different camera styles. \cite{zhu2020voc} trained a camera re-ID model to calculate a camera distance matrix. However, these methods are either only implemented on the target domain, or need to generate massive training data. And there are still few cameras in the current datasets. 

\section{GTA5 Synthetic Person Re-ID}\label{sec:dataset}
In this section, we first describe the data generation pipeline based on GTA5. As shown in Fig. \ref{data_overview}, the pipeline are mainly divided into four parts (\emph{i.e.,} location selection, attribute setting, person selection and data labeling). Then we construct a large-scale synthetic dataset and analyze its detailed properties. Finally, a mini dataset with ideal visual factors is introduced to expect to achieve excellent performance under the condition of delicate design and small data volume.

\subsection{Data Generation}

\ \quad \textbf{Location Selection. } Grand Theft Auto V (GTA5) is a computer game published in 2013. In GTA5, game developers construct a virtual world, which covers an area of 252 square kilometers and contains various visual factors close to real-world conditions. Besides, the game allows players to freely roam to explore abundant gameplay content. In the virtual world of GTA5, we selected 35 typical locations as the background of the synthetic person re-ID data, including mall, street, plaza, mountain, and so on.

\textbf{Attribute Setting. }
Following the scene selection, we need to set some essential parameters for each location. Firstly, inspired by \cite{sun2019dissecting}, since the different views of each person contain different nontrivial details, we set multiple viewpoints in each location, where each viewpoint contains multiple parameters with different values, such as location, height, and rotation angle. As a result, a total of 364 scenes are created. Then in order to obtain the controlled scenes, we first delimit a polygon area to place the person model, which is named “Region of Interest (ROI)”. The main purpose of this operation is to make each person appear in the desired area. Finally, to enhance diversity of the generated data, weather condition and time are randomly set to simulate the natural variation of illumination in real-world scenarios.

\textbf{Person Generation. } Since persons are the core of re-ID tasks, it is necessary to elaborately describe the person model in the data generation procedure. Firstly, due to the limitation of GTA5, the number of person categories provided by the game is less than 1,000. But we surprisingly find out that the publisher of GTA5 allows players to develop specific person models for non-commercial use to satisfy specific needs. Therefore, in order to create more person identities, we generate additional person models by collecting a large number of add-on person models and using some auxiliary software (\emph{i.e.,} openIV and addonpeds) to implement transformation from game models to synthetic persons, where the person models are collected from Internet and include many movie stars, sports players, historical political figures, \emph{etc}. For each person model, it contains up to 12 variations on the appearance, such as haircut, mask, and clothing. Overall, we adopt the 2369 person models, which have significant differences in skin color, gender, shape, clothes, \emph{etc}.

\begin{figure}[htbp]
	\centering
	\subfloat[Time stamp distribution.]{
		\begin{minipage}[t]{0.447\linewidth}
			\centering
			\label{fig2a}
			\includegraphics[width=1.0\columnwidth]{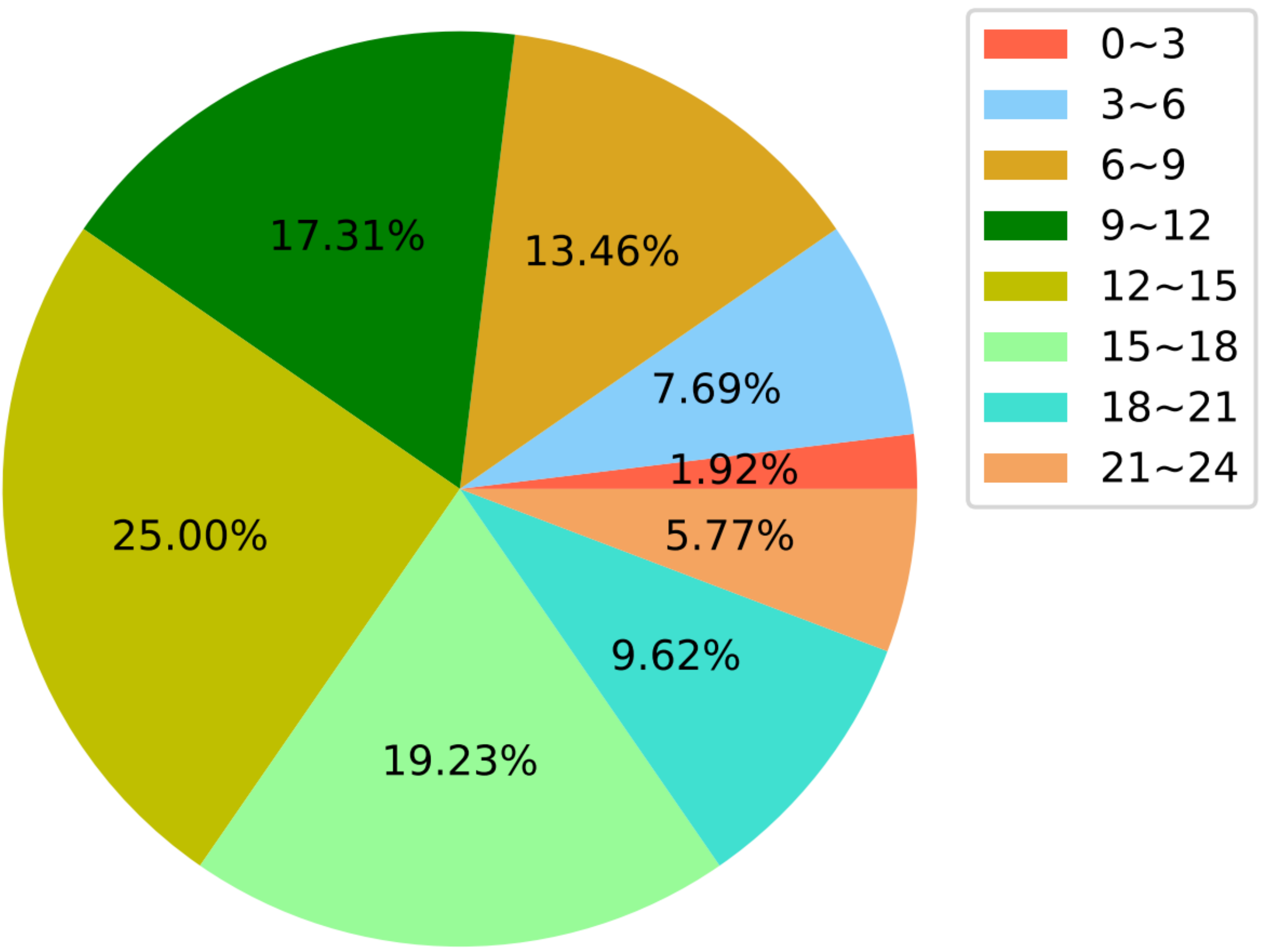}
		\end{minipage}%
	}%
	\subfloat[Weather condition distribution.]{
		\begin{minipage}[t]{0.50\linewidth}
			\centering
			\label{fig2b}
			\includegraphics[width=1.0\columnwidth]{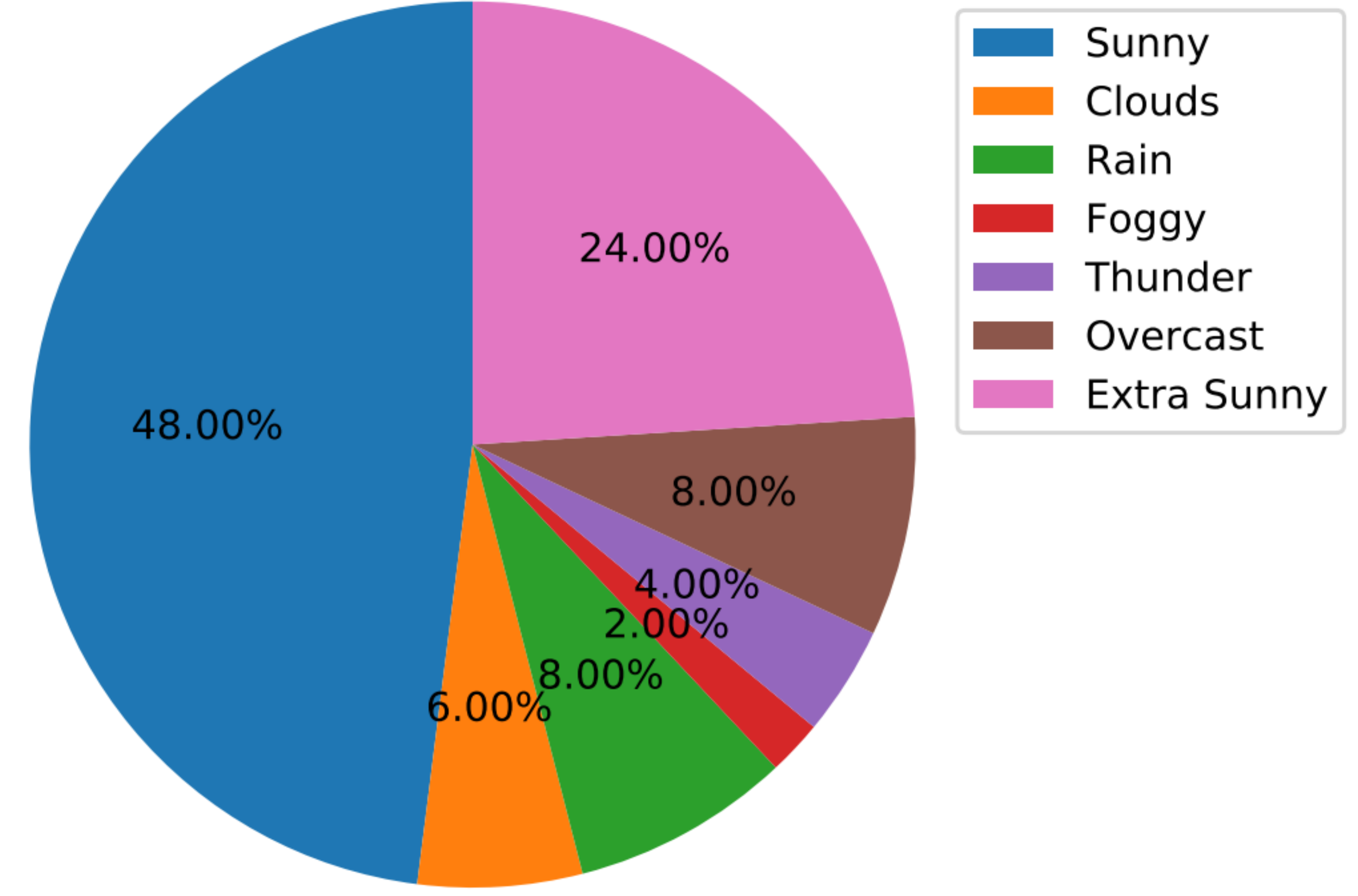}
		\end{minipage}%
	}%
	\centering 
	\caption{The pie charts with timestamp and weather condition distribution on our GSPR dataset. In the pie chart \ref{fig2a}, the label "0 $\sim$ 3" represents the time period of [0: 00, 3:00) within 24 hours a day.} \label{sector}
\end{figure}

Furthermore, during the data generation, each person walks along a scheduled route and does a random action in the chosen scene. We set multiple viewpoints in each position and each person has multiple distinctive traits. when human faces may not be easily observed in some viewpoints, it is feasible to recognize people from other biological characteristics or other viewpoints. Similar to the person re-ID in real-world scenarios, we are able to identify persons by their face, or other characteristics such as gender, age, clothes, and hairstyle from multiple viewpoints.

\textbf{Data Labeling. } The last step is to label the generated samples, where we utilize the hash value of objects provided by GTA5 to determine the identity of each person model. Due to the powerful ability of Mask R-CNN \cite{he2017mask} in semantic segmentation, we apply it to automatically detect persons in the generated samples and produce the corresponding bounding boxes. In addition, we also add the information of viewpoints and cameras (locations), which are set as the important parts of the identity label for each sample.

\subsection{Property Analysis for GSPR}
Based on the above data generation process, we first construct a large-scale GTA5 synthetic person re-id dataset, namely GSPR, which consists of 739,218 images with 2369 identities. As shown in Table 1, compared with the existing person re-ID datasets, GSPR is a more large-scale synthetic re-ID dataset, both in terms of data volumes and person identities. In addition to these advantages, GSPR is more diverse than existing datasets.

\textbf{Viewpoint Diversity. }
GSPR captures 312 viewpoints in the virtual world of GTA5, which are distributed in 26 locations with significant difference, including indoor scenes: bank, museum, pub, parking lot, \emph{etc}, and outdoor scenes: walking street, plaza, beach and so on.  There are 12 viewpoints for each position. Specifically, a person moves freely along a scheduled route, and the camera capture images sequentially from 12 angles at the each position, where all cameras have a high resolution of 1920 $\times$ 1080. Therefore, the entire GSPR dataset collects 2369 (identities) $\times$ 26 (cameras) $\times$ 12 (angles) = 739,218 images.

\begin{figure}
	\begin{center}
		\centering
		\includegraphics[width=0.80\linewidth]{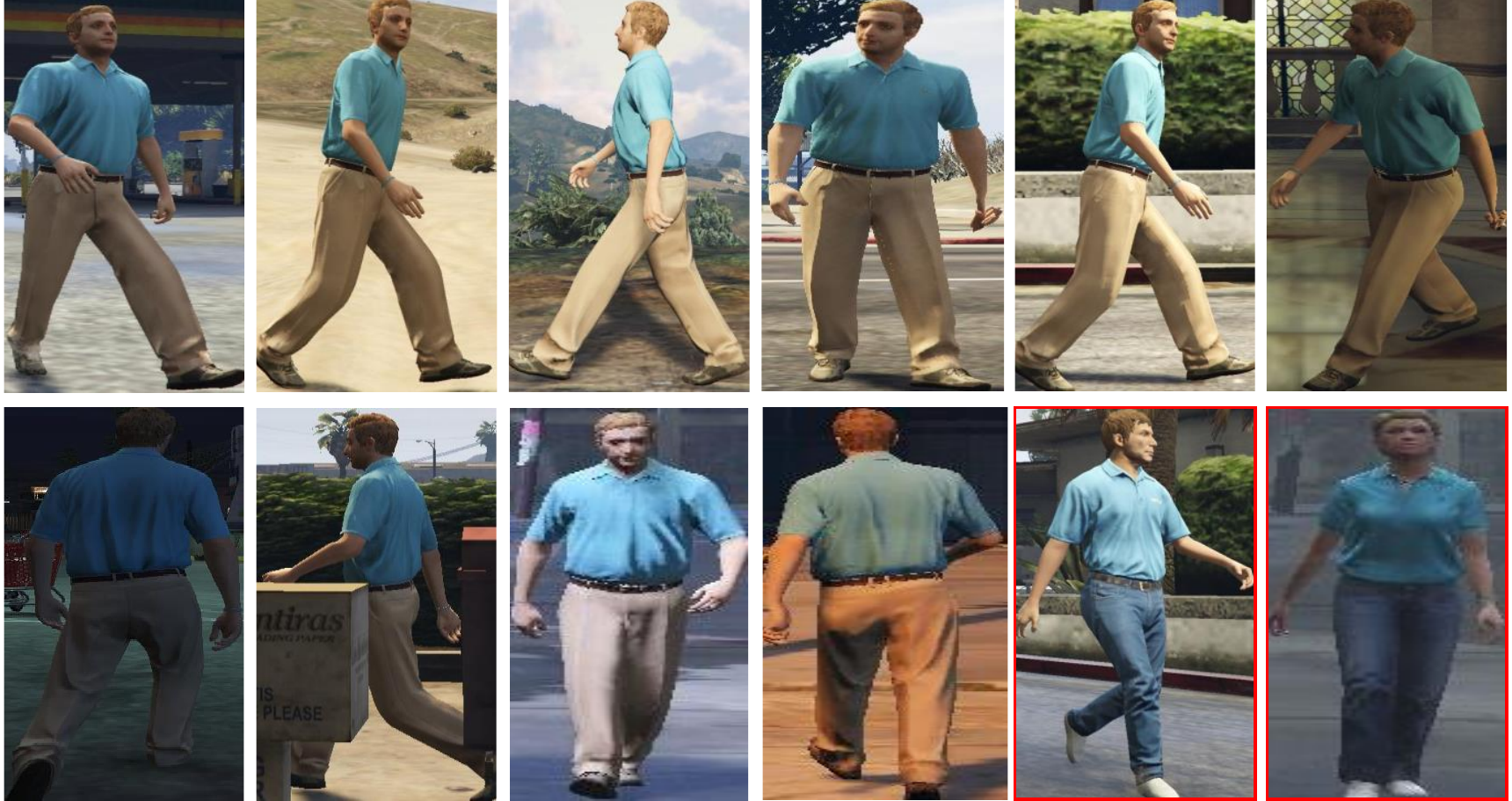}
		\hfill
	\end{center}
	\caption{Some examples in GSPR datatset. The first row represents samples in ideal situations. The second row represents samples in extreme situations, including illumination deficiency, occlusion, low image resolution, and similar but different people (i.e., the samples in red boxes).}\label{data_example}
\end{figure}

\textbf{Environment Diversity. } In order to generate the data that are close to the real-world scenarios, the images are captured at a random time in a day and under a random weather condition. In particular, we select common weathers, namely clear, clouds, rain, foggy, thunder, overcast and extra sunny. Two pie charts in Fig. \ref{sector} show the proportional distribution on the time stamp and weather conditions of the GSPR dataset, respectively. It can be found that although we tend to produce more images under common conditions (\emph{e.g.,} daytime and fine-weather scenes), some extreme cases are considered in our GSPR dataset, such as illumination deficiency, low resolution, and collision. Fig. \ref{data_example} also illustrate some examples in various cases.



\subsection{mini-GSPR}
Instead of considering various complex visual factors, we also construct a mini-scale synthetic person re-ID dataset under ideal conditions, referred as mini-GSPR. It consists of 27,040 images from 520 different identities. In mini-GSPR, we select 13 different locations (9 outdoor and 4 indoor scenes), and it is worth noting that there are only four overlapping locations between mini-GSPR and GSPR, including bank, parking lot, sidewalk street, and mountain area. In each location of mini-GSPR, the corresponding camera collects re-ID samples at four angles in sequence. Similar to GSPR, the resolution of cameras in mini-GSPR is 1920 $\times$ 1080, but mini-GSPR possesses the higher re-ID sample resolution, up to 948 $\times$ 522. Moreover, all samples in mini-GSPR are generated in the daytime scenes and sunny environment. There is no consideration of extreme cases (\emph{e.g.,} illumination deficiency and occlusion). Compared with GSPR and the existing datasets, mini-GSPR is a more small-scale person re-id dataset in the number of images and the number of identities. However, mini-GPR is still more diverse in the number of cameras than most datasets in terms of viewpoint and scene, except for MSMT17 and the proposed GSPR, and it has higher sample resolution and ideal visual factors that are difficult to provide in real scenarios. Overall, mini-GSPR dataset contains 520 (identities) $\times$ 13 (cameras) $\times$ 4 (angles) = 27,040 images.

\begin{figure*}
	\begin{center}
		\centering
		\includegraphics[width=0.95\linewidth]{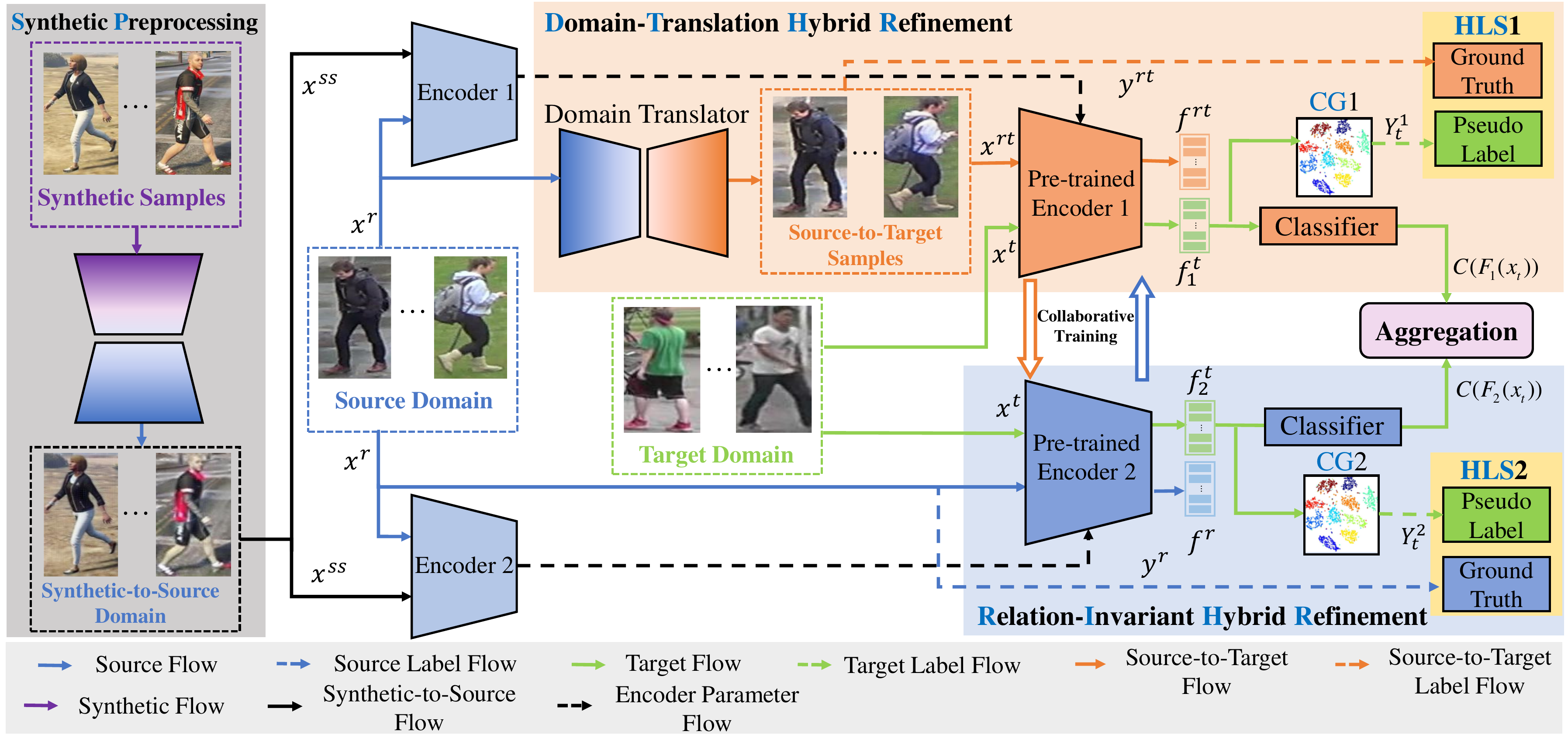}
		\hfill
	\end{center}
	\caption{Overview of our proposed Synthesis-based Multi-domain Collaborative Refinement (SMCR) Network. The overall architecture contains three modules: Synthetic Pretraining, Domain Translation Hybrid Refinement (DTHR), and Relation Invariant Hybrid Refinement (RIHR). \textbf{Synthetic Pretraining} module presents a new pretrained scheme to provide better initialization for the domain adaptive model. \textbf{DTHR} module integrates the processes of domain translation and pseudo label prediction, and performs joint training for the generated source-to-target samples and target domain data. \textbf{RIHR} module maintains original inter-sample relations invariant and implements synchronous learning for source domain and target domain data.}\label{overview}
\end{figure*}

\section{Proposed Method}\label{sec:method}
In this section, we propose a synthesis-based multi-domain collaborative refinement (SMCR) network for unsupervised cross-domain person re-ID. Specifically, we first give some essential symbolic definitions, and then introduce some components in the SMCR network, including a synthetic pretraining module and two complementary hybrid refinement modules (i.e., DTHR and RIHR). Finally, the proposed SMCR network is further refined by integrating the above modules and performing collaborative training between DTHR and RIHR to implement sufficient learning for the valuable knowledge from multiple domains. Fig. \ref{overview} illustrates the overview framework.

\subsection{Symbolic Definitions}
To facilitate understanding, some symbolic definitions are given here. Formally, we first assume access to synthetic data $D^S$ with $N_s$ labeled re-ID samples, where $D^S = \lbrace{(x_i^s, y_i^s)|^{N_s}_{i=1}}\rbrace$. Besides, given the real-world source domain data $D^R$ and target domain data $D^T$, $D^R = \lbrace{(x_i^r, y_i^r)|^{N_r}_{i=1}}\rbrace$ contains $N_r$ person images with corresponding ground-truth identity labels, and $D^T = \lbrace{(x_i^t, \hat{y}_i^t)|^{N_t}_{i=1}}\rbrace$ involves $N_t$ target images with pseudo labels predicted by the clustering algorithm. Furthermore, we denote source-to-target style-transferred samples as $D^{RT} = \lbrace{(x_i^{rt}, y_i^{rt})|^{N_{rt}}_{i=1}}\rbrace$, in which $x_i^{rt}$ and $y_i^{rt}$ are generated from the source domain data by a domain translator. Similarly, synthetic-to-source style transferred data with $D^{SR}$ can be represented as  $D^{SR} = \lbrace{(x_i^{sr}, y_i^{sr})|^{N_{sr}}_{i=1}}\rbrace$.

\subsection{Synthetic Pretraining}
Previous methods utilize the pretrained model based on ImageNet or synthetic datasets. However, the pretrained models in the single domain are not the best initialization for domain adaptive tasks. Besides, due to the severe shift between synthetic and real-world data, the model pre-trained on synthetic samples produces dramatic performance deterioration when generalized to real-world datasets.


To this end, a novel pretrained scheme is proposed to remedy the aforementioned problems. On the one hand, to reduce domain discrepancy between synthetic and real data, we translate the generated synthetic data to synthetic-to-source samples by resorting to the generative adversarial network, where the translated samples enjoy ground-truth identity labels derived from the synthetic data and have a similar style with the source domain. On the other hand, different from the previous methods that solely performed pretraining in a single domain, the designed model is pre-trained both on synthetic-to-source and source domain data, and we regard the pretraining process as performance generalization from synthetic-to-source samples to source domain. In this process, the parameters can be sufficiently trained to meet the needs of domain adaptive tasks in person re-ID, which is much better than traditional pretrained strategies. Finally, the pretrained model is utilized to domain adaptive learning between source domain and target domain.

\subsection{Domain Translation Hybrid Refinement}


Conventional UDA methods for person re-ID commonly aim to reduce the negative impacts of domain gap from two different perspectives, performing domain translation to generate source-to-target samples or predicting pseudo labels to refine finetuning in the target domain. However, as mentioned earlier, these methods have some limitations to different extent. To this end, we designed a domain translation hybrid refinement (DTHR) module to consider domain adaptation both in two aspects, which first integrates the processes of domain translation and pseudo label prediction, and then performs joint training with the source-to-target and target domain data during feature encoding. To be specific, as shown in Fig. \ref{overview}, there are three sub-modules in DTHR, domain translator $\mathcal{T}$, feature encoder $\mathcal{E}_1$, and pseudo-label generator $\mathcal{G}_1$. They are parameterized as $\theta_t$, $\theta_{e1}$ and $\theta_{g1}$, respectively.

\textbf{Domain Translator. }  The domain translator achieves feature translation from the annotated source domain to unlabeled target domain, and generates source-to-target samples with the same ground-truth labels as the source domain and a similar style to the target domain. The translator focuses on maintaining identity-related information invariant during translation, and it is based on the generator from a generative adversarial network. Given a real-world source data pair $(x^{r}, y^{r})$ as input, the corresponding source-to-target sample pair $(x^{rt}, y^{rt})$ is generated by the domain translator $\mathcal{T}$, which can be formally expressed as:

\begin{equation}
	\begin{split}
		\left \{
		\begin{array}{ll}
			x^{rt}_i = \mathcal{T}(x^r_i; \theta_t), \quad i \in [1, N_{rt}],  \\
			y^{rt}_i = y^r_i, \quad i \in [1, N_{rt}].
		\end{array}
		\right.
	\end{split}
\end{equation}
Because domain translation is not the focus of this paper, we utilize eSPGAN\cite{deng2018similarity} as our domain translator $\mathcal{T}$ in DTHR. The corresponding loss $\mathcal{L}_{dt}$ can be formulated as:
\begin{equation}
	\mathcal{L}_{dt} =  \mathcal{L}_{\mathcal{T}adv} + \mathcal{L}_{\mathcal{R}adv} + \mu_1 \mathcal{L}_{cyc} + \mu_2 \mathcal{L}_{ide} + \mu_3 \mathcal{L}_{c},
	\label{loss1}
\end{equation}
where the losses $\mathcal{L}_{\mathcal{T}adv}$ and $\mathcal{L}_{\mathcal{R}adv}$ indicate the standard adversarial losses for $D^R$ and $D^T$, $\mathcal{L}_{cyc}$ denotes the cycle consistency loss, $\mathcal{L}_{ide}$ is the inside-domain identity constraint to preserve the color composition during translation. $\mathcal{L}_{c}$ is the cross-entropy loss based on the feature learning, which is performed on source-to-target samples using ID-discriminative Embedding (IDE+). Moreover, $\mu_1$, $\mu_2$ and $\mu_3$ control the relative importance of the corresponding loss, respectively. We recommend referring the readers to CycleGAN \cite{zhu2017unpaired} and eSPGAN \cite{deng2018similarity} for more detailed information about loss functions. 



\textbf{Feature Encoder. } The feature encoder in DTHR is designed to perform joint training for the source-to-target and target domain data, where the extractor is progressively optimized by hybrid learning to provide more discriminative target representation during finetuning.
Formally, given a source-to-target image $x^{rt}_i$ and target domain image $x^{t}_i$, the corresponding output features $f^{t}_{1,i}$ and $f^{rt}_{i}$ of the encoder are obtained by the following mapping:

\begin{equation}
	\begin{split}
		\left \{
		\begin{array}{ll}
			f^{rt}_i = \mathcal{E}_1(x^{rt}_i; \theta_{e1}), \quad i \in [1, N_{rt}],  \\
			f^{t}_{1,i} = \mathcal{E}_1(x^{t}_i; \theta_{e1}), \quad i \in [1, N_{t}].
		\end{array}
		\right.
	\end{split}
	\label{3}
\end{equation}

Furthermore, in order to implement joint training for source-to-target and target data, we introduce a joint contrastive loss $\mathcal{L}_{joint}^1$. It can be defined as:

\begin{equation}
	\begin{split}
		\mathcal{L}_{joint}^1 =& - \frac{1}{N_{rt}+N_{t}}\bigg( \sum^{N_{rt}}_{i=1} log \Big(\frac{l^1(x^{rt}_i)}{\sum \limits^{M_{rt}}_{j=1}l^1_j(x^{rt}_i) + \sum \limits^{M_{t}^1}_{j=1}l^1_j(x^{t}_i)}\Big)   \\
		\cdot& y^{rt}_i + \sum \limits^{N_{t}}_{i=1} log \big(\frac{l^1(x^{t}_i)}{\sum \limits^{M_{rt}}_{j=1}l^1_j(x^{rt}_i) + \sum \limits^{M_{t}^1}_{j=1}l^1_j(x^{t}_i)}\big) \cdot y^{t}_{1,i} \bigg),
	\end{split}
	\label{4}
\end{equation}

\begin{equation}
	\begin{split}
		\left \{
		\begin{array}{ll}
			l^1_j(x^{rt}_i) = exp(<f^{rt}_i, y^{rt}_j>/\tau), \quad j \in [1, M_{rt}],  \\
			l^1_j(x^{t}_i) = exp(<f^{t}_{1,i}, y^{t}_j>/\tau), \quad j \in [1, M_{t}^1],
		\end{array}
		\right.
	\end{split}
	\label{5}
\end{equation}
where $<\cdot,\cdot>$ represents the similarity measure by the inner product for two input feature vectors. $\tau$ denotes a temperature hyper-parameter. $M_{rt}$ and $M_t^1$ are defined as the number of source-to-target sample classes and target domain pseudo categories in DTHR module. Overall, the encoded feature vector is encouraged to close its assigned classes using the above loss during joint training.

\textbf{Pseudo-label Generator. } To implement discriminative feature learning on target domain and remedy the adverse impact of noise pseudo labels, we first predict the pseudo labels for the target domain data using DBSCAN cluster algorithm as the pseudo-label generator $\mathcal{G}_1$ in DTHR, and then concatenate the predicted pseudo labels and the source ground-truth identity labels to construct a hybrid-label system HLS1. Finally, we refine the predicted pseudo labels by novel cluster criteria, which aims to preserve more reliable data in clusters by measuring the reliability of each feature point during clustering. Concretely, given a feature vector $f^{t}_{1,i}$ , our cluster criteria can be represented as:

\begin{equation}
	\begin{split}
		\left \{
		\begin{array}{ll}
			c_1(f^{t}_{1,i}) =max(0, \frac{|\gamma(f^{t}_{1,i}) \cap \gamma_s(f^{t}_{1,i})|}{|\gamma(f^{t}_{1,i}) \cup \gamma_s(f^{t}_{1,i})|} + \frac{|\gamma(f^{t}_{1,i}) \cap \gamma_e(f^{t}_{1,i})|}{|\gamma(f^{t}_{1,i}) \cup \gamma_e(f^{t}_{1,i})|} - \eta_1),   \\
			c_2(f^{t}_{1,i}) =max(0, \frac{|\gamma(f^{t}_{1,i}) \cap \gamma_s(f^{t}_{1,i})|}{|\gamma(f^{t}_{1,i}) \cup \gamma_s(f^{t}_{1,i})|} - \frac{|\gamma(f^{t}_{1,i}) \cap \gamma_e(f^{t}_{1,i})|}{|\gamma(f^{t}_{1,i}) \cup \gamma_e(f^{t}_{1,i})|} - \eta_2),
		\end{array}
		\right.
	\end{split}
	\label{6}
\end{equation}
where $\gamma(f^{t}_{1,i})$ denotes the cluster set containing the feature $f^{t}_{1,i}$,  $\gamma_s(f^{t}_{1,i})$ and $\gamma_e(f^{t}_{1,i})$ are the clusters including $f^{t}_{1,i}$ when shrinking and enlarging the distance threshold in the DBSCAN algorithm, respectively.
Two dynamic thresholds (\emph{i.e.} $\eta_1$ and $\eta_2$) and two corresponding clutering criteria (\emph{i.e.} $c_1(f^{t}_{1,i})$ and $c_2(f^{t}_{1,i})$) are obtained by identifying reliable feature points in the clusters on-the-fly. Overall, we analyze the reliability of feature points from target domain data by comparing their clustering results with the results when shrinking and enlarging the threshold distance, respectively. Finally, the corresponding pseudo-label $\hat y_{1,i}^{t}$ can be obtained by the following mapping:
\begin{equation}
	\begin{split}
		\hat{y}^{t}_{1,i} = \left \{
		\begin{array}{ll}
			\mathcal{G}_1(f^{t}_{1,i}; \theta_{g1}),  &c_1(f^{t}_{1,i}) > 0, c_2(f^{t}_{1,i}) > 0,\\
			-1.  &otherwise,
		\end{array}
		\right.
	\end{split}
	\label{7}
\end{equation}
when $\hat{y}_{1,i}^{t} = -1$, the corresponding sample is regarded as discrete instances with individual class labels. Therefore, the number of pseudo labels equals $M^t_c+M^t_o$, where $M^t_c$ and $M^t_o$ represent the number of reliable clusters and discrete instances, respectively.

\subsection{Relation Invariant Hybrid Refinement}
Although source-to-target data enjoys the similar style to the target domain and remedy the adverse impact of the domain discrepancy, it has been found in \cite {ge2020structured} that the generated data loses the original inter-sample relations from the source domain after translation.

To this end, we devise a relation-invariant hybrid refinement (RIHR) architecture, which aims to preserve the original inter-sample relations and perform joint learning for source domain and target domain data. As shown in Fig. \ref{overview}, there is a feature encoder $\mathcal{E}_2$ and a pseudo-label generator $\mathcal{G}_2$ in our RIHR module, which have the same structural details as sub-modules in DTHR, but the parameters of these sub-modules are not shared between DTHR and RIHR. Furthermore, two sub-modules in RIHR are parameterized with $\theta_{e2}$ and $\theta_{g2}$, respectively, and we generate a new hybrid-label system HLS2 by concatenating identity labels from the source domain and the new target pseudo label set. These parameterized sub-modules are iteratively trained to obtain the discriminative features in the target domain, where we simultaneously utilize source domain data with inter-sample relations and pseudo-labeled target domain data. Formally, the output feature of the encoder in RIHR can be formulated as:
\begin{equation}
	\begin{split}
		\left \{
		\begin{array}{ll}
			f^{r}_i = \mathcal{E}_2(x^{r}_i; \theta_{e2}),  & i \in [1, N_r] \\
			f^{t}_{2,i} = \mathcal{E}_2(x^{t}_i; \theta_{e2}),  & i \in [1, N_t].
		\end{array}
		\right.
	\end{split}
	\label{8}
\end{equation}
The joint contractive loss $\mathcal{L}_{joint}^2$ is written as:

\begin{equation}
	\begin{split}
		\left \{
		\begin{array}{ll}
			l^2_j(x^{r}_i) = exp(<f^{r}_i, y^{r}_j>/\tau), \quad j \in [1, M^{R}],  \\
			l^2_j(x^{t}_i) = exp(<f^{t}_{2,i}, y^{t}_j>/\tau), \quad j \in [1, M^{T}_2],
		\end{array}
		\right.
	\end{split}
	\label{9}
\end{equation}

\begin{equation}
	\begin{split}
		\mathcal{L}_{joint}^2 =& - \frac{1}{N_{r}+N_t}\bigg( \sum \limits^{N_{r}}_{i=1} log \Big(\frac{l^2(x^{r}_i)}{\sum \limits^{M_{r}}_{j=1}l^2_j(x^{rt}_i) + \sum \limits^{M_{t}^2}_{j=1}l^2_j(x^{t}_i)}\Big)  \\
		\cdot& y^{r}_i + \sum \limits^{N_{t}}_{i=1} log \big(\frac{l^2(x^{t}_i)}{\sum \limits^{M_{t}}_{j=1}l^2_j(x^{rt}_i) + \sum \limits^{M_{t}^2}_{j=1}l^2_j(x^{t}_i)}\big) \cdot y^{t}_{2,i} \bigg).
	\end{split}
	\label{10}
\end{equation}


\subsection{SMCR Network}
In order to implement synthesis-based multiple domains collaborative training and refinement, we construct a novel network by assembling the devised synthetic pretraining, DTHR and RIHR modules, referred as SMCR network. We first utilize the abundant information from synthetic samples, source domain and target domain as the input of the SMCR network for jointly learning valuable knowledge from multiple domains. To be specific, for synthetic samples, a novel pretrained scheme is used in the synthetic pretraining module, which provides better initialization for subsequent modules by combining with source domain data; for source domain data, we aim to reduce the negative impact of domain gaps by generating style-transferred samples in DTHR, and the original inter-sample relations invariant is considered by directly learning the source rich information in RIHR; for target domain data, the valuable complementary knowledge is obtained from two branches (\emph{i.e.} DTHR and RIHR) using predicting pseudo labels based on our cluster criteria.

Furthermore, to further remedy the adverse side of the noisy pseudo-labels and optimize the discrimination of target features, we introduce soft pseudo-labels by performing collaborative refinement between the DTHR module and the RIHR module in a mutual-learning manner. To prevent the two modules from converging so close to each other that their output independence is lost, inspired by MoCo \cite{he2020momentum} and MMT \cite{ge2020mutual}, we construct a momentum update encoder for each module to extract discriminative target features as additional soft pseudo-labels for supervising another module. Formally, the parameters of momentum update encoders from DTHR module and RIHR module at \emph{k-th} iteration  can be indicated as $A^k(\theta_{e1})$ and $A^k(\theta_{e2})$, respectively. They are able to be obtained by:
\begin{equation}
	\begin{split}
		\left \{
		\begin{array}{ll}
			A^k(\theta_{e1}) = \lambda A^{k-1}(\theta_{e1}) + (1-\lambda)\theta_{e1},  \\
			A^k(\theta_{e2}) = \lambda A^{k-1}(\theta_{e2}) + (1-\lambda)\theta_{e2},
		\end{array}
		\right.
	\end{split}
	\label{11}
\end{equation}
where $A^{k-1}(\theta_{e1})$ and $A^{k-1}(\theta_{e2})$ are the parameters of  momentum update encoders from the two modules at iteration $k-1$, $\theta_{e1}$ and $\theta_{e2}$ are updated by back-propagation in each iteration. It is worth noting that we set $A^{0}(\theta_{e1}) = \theta_{e1}$ and $A^{0}(\theta_{e2}) = \theta_{e2}$ in first iteration. $\lambda \in \left[0,1\right)$ is a momentum coefficient. Furthermore, we construct a collaborative triple loss between the two modules (\emph{i.e.,} DTHR and RIHR). Given a target domain image $x^t_i$, the collaborative triple loss $L_{col}$ is able to be formulated as:
\begin{equation}
	L_{col} = \alpha L^1_{col} + (1-\alpha) L^2_{col},
	\label{12}
\end{equation}
where $L^1_{col}$ indicates the collaborative training loss of RIHR’s momentum update encoder for the DTHR module and $L^2_{col}$ is vise versa. The hyper-parameter $\alpha$ controls the contribution from each loss term:
\begin{equation}
	\begin{split}
		\left \{
		\begin{array}{ll}
			L^1_{col} = -\frac{1}{N^T}\sum \limits_{i=1}^{N^T} [S(x^{t}_i, \theta_{e1}) \cdot \log{S(x^{t}_i, A^k(\theta_{e2}))}],  \\
			L^2_{col} = -\frac{1}{N^T}\sum \limits_{i=1}^{N^T} [S(x^{t}_i, \theta_{e2}) \cdot \log{S(x^{t}_i, A^k(\theta_{e1}))}],
		\end{array}
		\right.
	\end{split}
	\label{13}
\end{equation}
\begin{equation}
	S(x^t_i, \theta_{e1}) = \frac{exp(D^{+}(x^t_i, \theta_{e1} ))}{exp(D^{+}(x^t_i, \theta_{e1}) +D^{-}(x^t_i, \theta_{e1})))},
	\label{14}
\end{equation}
where $S(x^t_i, \theta_{e1})$ represents the softmax triple item generated by DTHR module, and its correponding soft label is $S(x^{t}_i, A^k(\theta_{e2}))$ from RIHR module. Furthermore,

\begin{equation}
	\begin{split}
		\left \{
		\begin{array}{ll}
			D^{+}(x^t_i, \theta_{e1} ) = ||\mathcal{E}_1(x^{t}_{i}; \theta_{e1})-\mathcal{E}_1(x^{t}_{i,+}; \theta_{e1})||,  \\
			D^{-}(x^t_i, \theta_{e1} ) = ||\mathcal{E}_1(x^{t}_{i}; \theta_{e1})-\mathcal{E}_1(x^{t}_{i,-}; \theta_{e1})||,
		\end{array}
		\right.
	\end{split}
	\label{15}
\end{equation}
where $x^{t}_{i,-}$ and $x^{t}_{i,+}$ inicate the hardest positive and negative samples for the input $x^t_i$ in a mini-batch. $D^{-}(x^t_i, \theta_{e1})$ and $D^{+}(x^t_i, \theta_{e1})$ are the Euclidean distance between the feature of $x^t_i$ and the features of $x^{t}_{i,-}$ and $x^{t}_{i,+}$, repectively. 

At the last stage of SMCR, we perform a weighted average for the learned information from two branches by weighted factor $\alpha$ to further refine the discriminative ability of the model, and the averaging results are regarded as the final outputs of the SMCR network. Therefore, given target domian image $x^{t}_i$, the corresponding final output $y_i$ of SMCR network is:

\begin{equation}
	\begin{aligned}
		y_i &= \alpha\mathcal{C}_1(f^t_{1,i}) + (1 - \alpha)\mathcal{C}_2(f^t_{2,i}) \\
		&= \alpha\mathcal{C}_1(\mathcal{E}_1(x^{t}_i; \theta_{e1})) + (1 - \alpha)\mathcal{C}_2(\mathcal{E}_2(x^{t}_i; \theta_{e2})),\\
	\end{aligned}
	\label{16}
\end{equation} 
where $\mathcal{C}_1$ and $\mathcal{C}_2$ are two learnable classifiers from DTHR and RIHR to predict person identity label.

The overall loss function combines the joint training losses (\emph{i.e., $\mathcal{L}_{joint}^1$ and $\mathcal{L}_{joint}^2$} ) from the DTHR module and the RIHR module with the collaborative triple loss using hyperparameters $\alpha$ and $\beta$, which can be defined as:
\begin{equation}
	\mathcal{L} = \beta L_{col} + 2(1-\beta) (\alpha \mathcal{L}_{joint}^1 + (1-\alpha) \mathcal{L}_{joint}^2).
	\label{17} 
\end{equation}

\begin{table*}[h]
	\setlength{\abovecaptionskip}{0.cm}
	\begin{center}
		\begin{spacing}{1.13}
			\caption{ Comparison of our SMCR network with state-of-the-art methods on unsupervides domain adaptive person re-ID. We use bold fonts to highlight he best results.}\label{table4}
			\resizebox{1.0\hsize}{!}{
				\begin{tabular}{c| c|c c c c|c c c c}
					\hline
					\multicolumn{2}{c|}{\multirow{2}{*}{Methods}}  & \multicolumn{4}{c|}{DukeMTMC-reID $\rightarrow$ Market-1501}  & \multicolumn{4}{c}{Market-1501 $\rightarrow$ DukeMTMC-reID}  \\
					\cline{3-10} 
					\multicolumn{2}{c|}{}   & {mAP} & {Rank-1} & {Rank-5}  & {Rank-10}  & {mAP} & {Rank-1} & {Rank-5}  & {Rank-10} \\
					\hline

					SSG \cite{fu2019self} &ICCV'19 &58.3 & 80.0 &90.0 &92.4 &53.4 & 73.0 &80.6 & 83.2  \\
					
					MMCL \cite{wang2020unsupervised} &CVPR'20 &60.4 &84.4 &92.8 &95.0 &51.4 &72.4 &82.9 &85.5\\
					ECN++ \cite{zhong2020learning} &TPAMI'20 &63.8 & 84.1 & 92.8 &95.4 &54.4 & 74.0 &83.7 &87.4 \\
					DG-Net++ \cite{zou2020joint} &ECCV'20 &63.8 &78.9 &87.8 &90.4 &61.7 &82.1 &90.2 &92.7 \\
					JVTC+ \cite{li2020joint}  &ECCV'20 & 67.2 &86.8 &95.2 &97.1 &66.5 &80.4 &89.9 &92.2\\
					GPR \cite{luo2020generalizing}  &ECCV'20 &71.5 &88.1 &94.4 &96.2 &65.2 &79.5 &88.3 &91.4 \\
					SDA \cite{ge2020structured}   &arXiv'20 &74.3 &89.7 &95.9 &97.4 &66.7 &79.9 &89.1 &92.7 \\
					MMT \cite{ge2020mutual}   &ICLR'20  &76.5 &90.9 &96.4 &97.9 &67.3 &80.8 &90.3 &93.0 \\
					SPCL \cite{ge2020self} &NeurIPS'20 &79.2  &91.5 & 96.9 &98.0 &69.9 &83.4 &91.0 &93.1\\ 
					GLT \cite{zheng2021group} &CVPR'21 &79.5 &92.2 &96.5 &97.8 &69.2 &82.0 &90.2 &92.8 \\
					ABMT \cite{chen2021enhancing} &WACV'21 &80.4 &93.0 &- &- &70.8 &83.3  &- &-  \\ \hline
					\multicolumn{2}{c|}{SMCR  (\emph{mini-GSPR})} &82.4 &92.8 &97.1 &\textbf{98.3} &73.2 &\textbf{85.3} &91.9 &94.1 \\
					\multicolumn{2}{c|}{\textbf{SMCR}  (\emph{GSPR})} &\textbf{83.0} &\textbf{93.5} &\textbf{97.2} &\textbf{98.3} &\textbf{73.7} &84.8 &\textbf{93.1} &\textbf{94.7} \\ \hline 
					\multicolumn{10}{c}{}
				\end{tabular}\label{comparasion}
			}
			\resizebox{1.0\hsize}{!}{
				\begin{tabular}{c| c|c c c c|c c c c}
					\hline
					\multicolumn{2}{c|}{\multirow{2}{*}{Methods}}  & \multicolumn{4}{c|}{Market-1501 $\rightarrow$ MSMT17}  & \multicolumn{4}{c}{DukeMTMC-reID $\rightarrow$ MSMT17}  \\
					\cline{3-10} 
					\multicolumn{2}{c|}{}   & {mAP} & {Rank-1} & {Rank-5}  & {Rank-10}  & {mAP} & {Rank-1} & {Rank-5}  & {Rank-10} \\
					\hline

					SSG \cite{fu2019self} &ICCV'19 &8.5 &25.3 &36.3 &42.1 &13.3 &32.2 &- & 51.2   \\
					ECN+ \cite{zhong2020learning} &TPAMI'20 &15.2 &40.4 &53.1 &58.7 &16.0 & 42.5 &55.9 &61.5 \\
					JVTC+ \cite{li2020joint}  &ECCV'20 &25.1 &48.6 &65.3 &68.2 &27.5 &52.9 &70.5 &75.9 \\
					SDA \cite{ge2020structured}   &arXiv'20 &23.2 &49.5 &62.2 &67.7 &30.3 &59.6 & 71.7 &76.2 \\
					GLT \cite{zheng2021group} &CVPR'21 &26.5 &56.6 &67.5 &72.0 &27.7 &59.5 &70.1 &74.2 \\
					MMT \cite{ge2020mutual}   &ICLR'20   &26.6 &54.4 &67.6 &72.9 &29.7 & 58.8 &71.0 &76.1 \\
					SPCL \cite{ge2020self} &NeurIPS'20 &26.8 &53.7 &65.0 &69.8 &31.8 & 58.9 &70.4 &75.2 \\ 
					ABMT \cite{chen2021enhancing} &WACV'21 &27.8 &55.5 &- &- &33.0 &61.8  &- &-  \\ \hline
					\multicolumn{2}{c|}{SMCR  (\emph{mini-GSPR})} &32.7 &57.8 &70.6 &75.2 &33.9 &60.4 &72.8 &77.3 \\
					\multicolumn{2}{c|}{\textbf{SMCR}  (\emph{GSPR})}  &\textbf{32.8} &\textbf{58.3} &\textbf{70.7} &\textbf{75.7} &\textbf{34.7} &\textbf{64.9} &\textbf{74.8} &\textbf{78.3} \\ \hline
				\end{tabular}\label{comparasion}
			}
		\end{spacing}
	\end{center}
\end{table*}

\section{Experiments}\label{sec:experiment}
In this section, we first introduce the experimental datasets and implementation details. Then the experimental results of the proposed SMCR network on four domain adaptive person re-ID tasks are shown, and we compare them with state-of-the-art methods. Finally, the effectiveness of all the components is examined using the ablation study.

\subsection{Datasets and Implementation Details}
\textbf{Datasets. } The proposed approach is evaluated on five person re-ID datasets, including the proposed GSPR and mini-GSPR, real-world datasets Market-1501 \cite{zheng2015scalable}, DukeMTMC-reID \cite{ristani2016performance} and MSMT17 \cite{wei2018person}. The Market-1501 \cite{zheng2015scalable} dataset consists of 32,668 annotated images of 1,501 identities shot from 6 cameras in total, for which 12,936 images of 751 identities are used for training and 19,732 images of 750 identities are in the testset. DukeMTMC-reID \cite{ristani2016performance} contains 16,522 person images of 702 identities for training, and the remaining images out of another 702 identities for testing, where all images are collected from 8 cameras. Meanwhile, DukeMTMC-reID considers additional 408 identities as distractors. MSMT17 \cite{wei2018person} is the most challenging and large-scale real-world dataset consisting of 126,441 bounding boxes of 4,101 identities taken by 15 cameras, for which 32,621 images of 1,041 identities are spitted for training. But it is worth noting that MSMT is collected from 180 hours of high-resolution videos and three labelers annotated it for two months. Such costs are unaffordable in real-world application deployment. Comparatively, GSPR only spent 156 CPU-hours and mini-GSPR took 52 CPU-hours.

\textbf{Implementation Details. }
We adopt the IBN-ResNet50 as the backbone of our feature encoders (\emph{i.e.} $\mathcal{E}_1$ and $\mathcal{E}_2$), which is proposed by \cite{pan2018two} and aims to strengthen generalization capicity of deep neural networks for domain adaptative tasks. The training parameters include 50 epochs, batch size 64, momentum 0.2, and weight decay 0.0005. We initialize the learning rate to be 0.00035 and decrease it by 10 every 20 epochs. For two dynamic thresholds (\emph{i.e.} $\eta_1$ and $\eta_2$) in cluster criterion, they are set to the sum and difference of the top-90\% $\frac{|\gamma(f^{t}) \cap \gamma_s(f^{t})|}{|\gamma(f^{t}) \cup \gamma_s(f^{t})|}$ in the first epoch and the maximum $\frac{|\gamma(f^{t}) \cap \gamma_e(f^{t})|}{|\gamma(f^{t}) \cup \gamma_e(f^{t})|}$ in the cluster of each epoch. The hyperparameters $\lambda$ and $\beta$ in our SMCR network are set to 0.5 and 0.01, respectively. Besides, we adopt mean average precision (mAP) and CMC Rank-1, Rank-5, Rank-10 accuracies to evaluate the performance of models. It is worth noting that the predicted pseudo labels for the fine-tuning are updated after each epoch, so the mini-batch of target-domain images have to be re-organized with updated pseudo labels after each epoch. All images are resized to 256 $\times$ 128 when they are ready to feed into the networks. We conduct all experiments on 4 GTX-1080TI GPUs, and there are no post-processing technologies (\emph{e.g.} re-ranking and multi-query fusion) in our experiments.

\subsection{Comparison with State-of-the-arts.}
We compare our proposed approach with state-of-the-art UDA methods on four unsupervised cross-domain re-ID tasks in Table \ref{comparasion}. In order to illustrate the effectiveness of the synthetic pretraining module in the proposed method, we use the constructed synthetic datasets GSPR and mini-GSPR to participate in model pretraining, and the pretrained models are called “SMCR (GSPR)” and “SMCR (mini-GSPR)”, respectively. It can be seen in Table \ref{comparasion} that our SCMR network implements significantly performance improvements when comparing with state-of-the-art methods in terms of mAP and CMC Rank-1, Rank-5, Rank-10 accuracies. 

Specifically, we first verify the performance of our SCMR network by comparing it with domain translation-based approaches. SDA \cite{ge2020structured} used an online relationship consistency regularization term in a compact model to perform domain translation and maintain inter-sample relations invariant, which dominate state-of-the-art performance in domain translation-based methods. However, the proposed SMCR network surpasses SDA by a big margin. SCMR (mini-GSPR) obtains relative improvements of 8.1\% (82.4\% vs 74.3\%), 6.5\% (73.2\% vs 66.7\%), 9.5\% (32.7\% vs 23.2\%) mAP and 3.6\% (33.9\% vs 30.3\%) mAP on Market-1501 $\rightarrow$ DukeMTMC-reID, Market-1501 $\rightarrow$ DukeMTMC-reID, Market-1501 $\rightarrow$ MSMT17 and DukeMTMC-reID $\rightarrow$ MSMT17 domain adaptive tasks, respectively. SMCR (GSPR) gains further performance improvements against SDA by 8.7\%, 7.0\%, 9.6\%, and 4.4\% mAP on the four domain adaptive tasks. These results demonstrate the effectiveness of our SMCR network to learn valuable domain adaptive knowledge by enhancing domain translation in DTHR and maintaining inter-sample relations in RIHR.

We then evaluate the performance of the SMCR network and report the comparison results about pseudo-label-based methods. Due to remarkable performance, pseudo-label-based methods have been a hot research topic in the field of UDA for person re-ID. Recently, ABMT \cite{chen2021enhancing} gains optimal performance by using an asymmetric structure inside teacher-student networks. Comparing with ABMT, SMCR (mini-GSPR) still implements performance improvements by 2\% (82.4\% vs 80.4\%), 2.4\% (73.2\% vs 70.8\%) and 4.9\% (32.7\% vs 27.8\%), and 0.9\% (33.9\% vs 33.0\%) mAP on four domain adaptation tasks. Furthermore, the performance of SMCR (GSPR) yields new records on the unsupervised cross-domain person re-ID problem. These results show that the proposed method is able to provide more discriminative feature learning capacity using our cluster criteria and collaborative refinement. 

\begin{table}[h]
	\setlength{\abovecaptionskip}{0.cm}
	\begin{center}
		\begin{spacing}{1.3}
			\caption{Performance of our SMCR network for UDA tasks from synthetic datasets to real-world datasets.}\label{table4}
			\resizebox{1.0\hsize}{!}{
				\begin{tabular}{c|c|c c c c}
					\hline
					Source domain &Target domain   & {mAP} & {Rank-1} & {Rank-5}  & {Rank-10}\\
					\hline
					\multirow{3}{*}{mini GSPR}
					&Market-1501 &80.9 &91.3 &96.4 &97.7 \\
					&DukeMTMC-ReID &71.2 &83.4 &91.2 &93.2 \\
					&MSMT17 &34.2 &61.2 &73.0 &77.4 \\ \hline
					
					\multirow{3}{*}{GSPR}
					&Market-1501 &81.8 &92.2 &96.7 &97.7 \\
					&DukeMTMC-ReID &72.6 &84.2 &91.0 &93.5 \\
					&MSMT17 &36.1 &62.3 &73.9 &78.0 \\
					
					\hline
					
				\end{tabular}\label{uda_dataset}
			}
			
		\end{spacing}
	\end{center}	
\end{table}

\begin{table}[h]
	\setlength{\abovecaptionskip}{0.cm}
	\begin{center}
		\begin{spacing}{1.2}
			\caption{Ablation study of our proposed SMCR network.}\label{table5}
			\resizebox{1.0\hsize}{!}{
				\begin{tabular}{c|c c c c}
					\hline
					\multirow{2}{*}{Methods}  & \multicolumn{4}{c}{DukeMTMC-reID $\rightarrow$ Market-1501} \\
					\cline{2-5} 
					& {mAP} & {Rank-1} & {Rank-5}  & {Rank-10} \\
					\hline
					Baseline &32.9 &61.7 &77.1 &82.0 \\ \hline
					DTHR (\emph{ind}) &77.0 &90.5 &96.1 &97.4 \\
					RIHR (\emph{ind}) &80.6 &92.0 &97.1 &98.2 \\
					DTHR (\emph{col})  &81.3 &82.2 &92.7 &97.2 \\
					RIHR (\emph{col}) &82.2 &92.7 &97.2 &98.0  \\ \hline
					SMCR (\emph{w/o criteria})  &47.5 &70.2 &85.1 &89.4 \\
					SMCR (\emph{ResNet50})  &77.8 &90.9 &96.2 &97.6 \\
					SMCR (\emph{CycleGAN})  &80.3 &91.4 &96.8 &98.2 \\ \hline
					SMCR (\emph{w/o syn}) &81.8 &91.9 &97.0 &98.0 \\
					SMCR (\emph{GSPR-pre})  &81.0 &92.1 &96.7 &98.1 \\ 
					SMCR (\emph{mini-GSPR}) &82.4 &92.8 &97.1 &\textbf{98.3} \\
					SMCR (\emph{GSPR}) &\textbf{83.0} &\textbf{93.5} &\textbf{97.2} &\textbf{98.3} \\ \hline
					
				\end{tabular}\label{ablation}
			}
		\end{spacing}
	\end{center}
\end{table}

\begin{figure*}[htbp]
	\centering
	\subfloat[Impact of $\alpha$ for mAP]{
		\begin{minipage}[t]{0.25\linewidth}
			\centering
			\label{fig5a}
			\includegraphics[width=1.0\columnwidth]{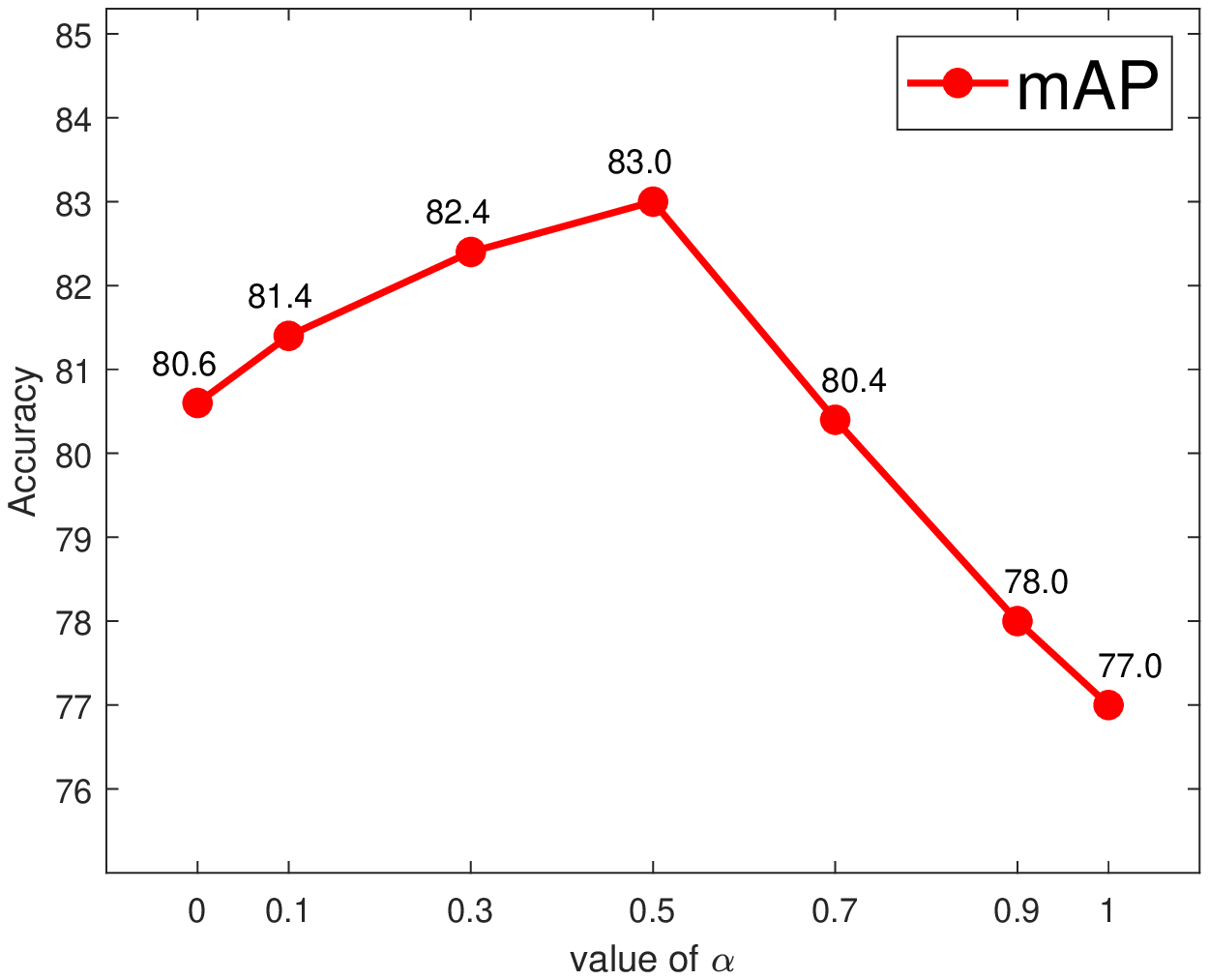}
		\end{minipage}%
	}%
	\subfloat[Imapct of $\alpha$ for Rank-1 accuracy]{
		\begin{minipage}[t]{0.25\linewidth}
			\centering
			\label{fig5b}
			\includegraphics[width=1.0\columnwidth]{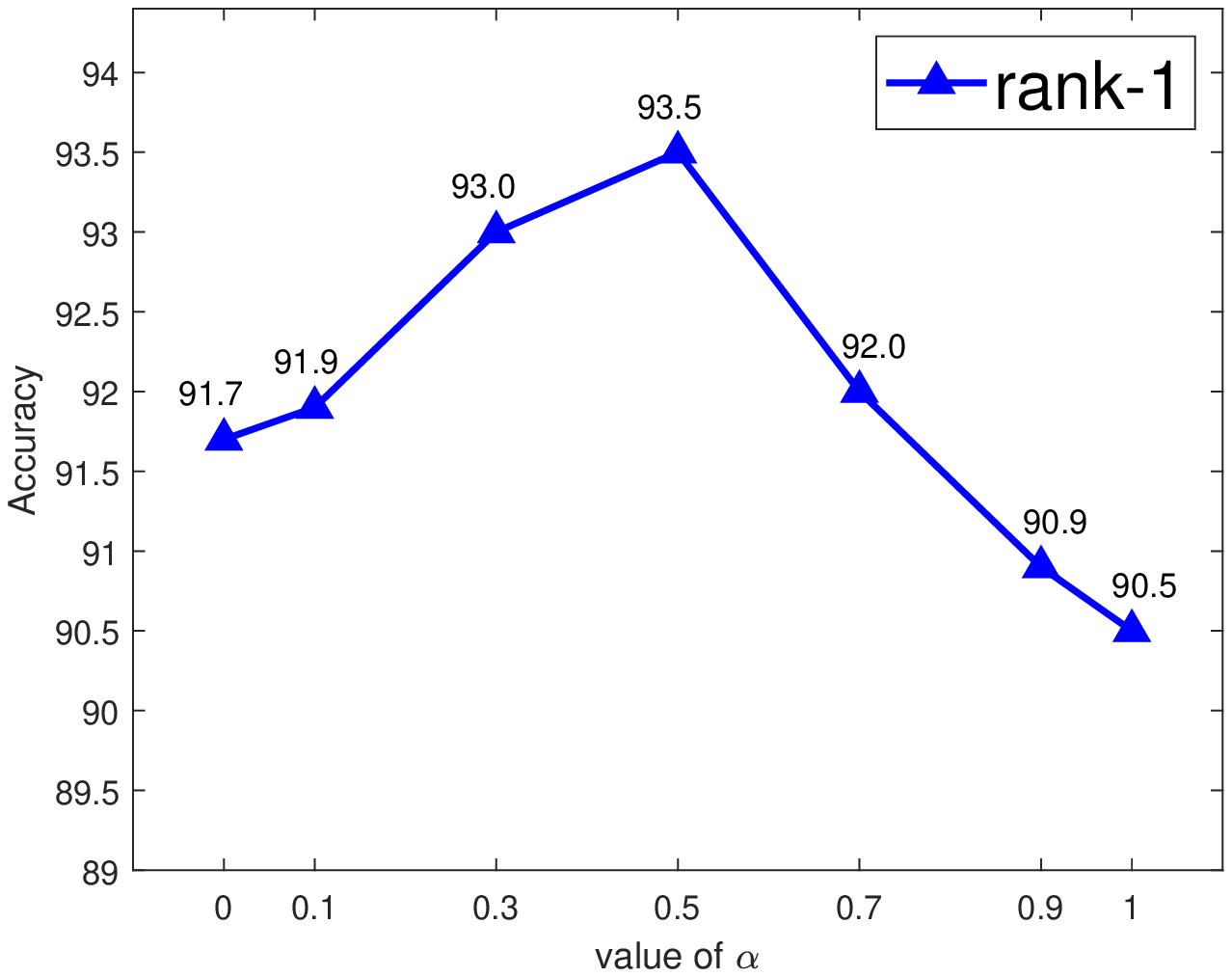}
		\end{minipage}%
	}%
	\subfloat[Imapct of $\beta$ for mAP accuracy]{
		\begin{minipage}[t]{0.25\linewidth}
			\centering
			\label{fig5c}
			\includegraphics[width=1.0\columnwidth]{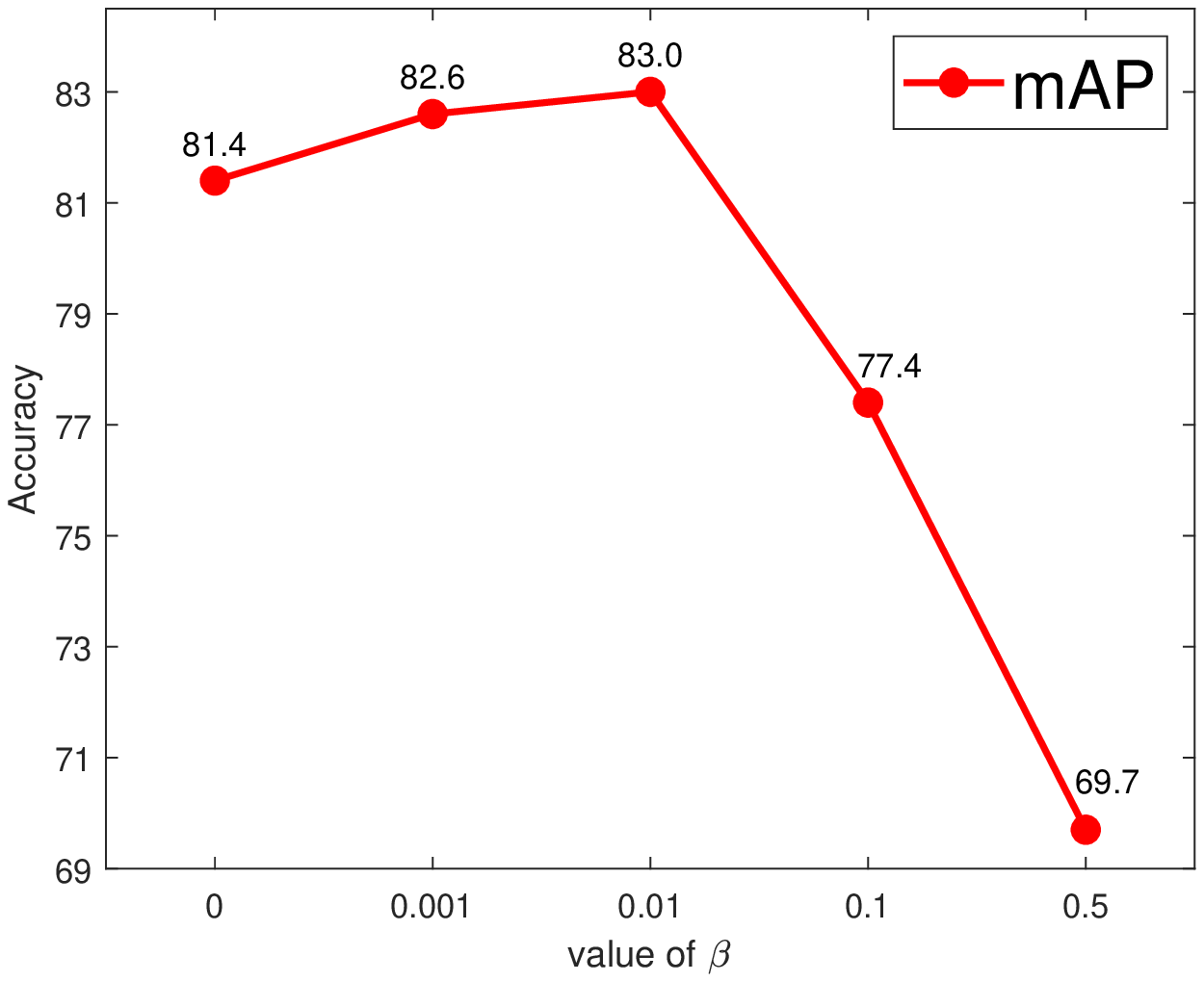}
		\end{minipage}%
	}%
	\subfloat[Imapct of $\beta$ for Rank-1 accuracy]{
		\begin{minipage}[t]{0.25\linewidth}
			\centering
			\label{fig5d}
			\includegraphics[width=1.0\columnwidth]{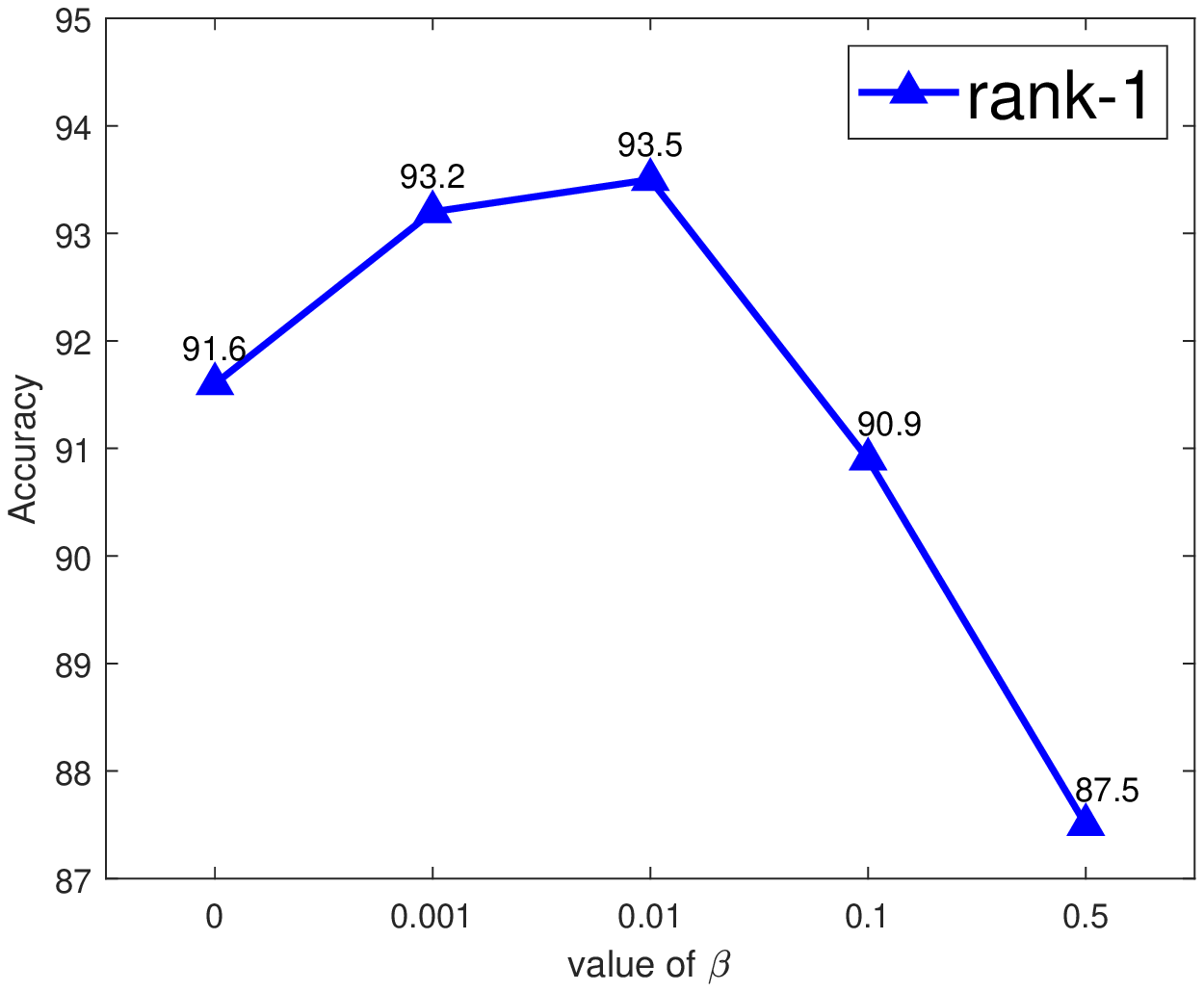}
		\end{minipage}%
	}%
	
	\centering
	\subfloat[Imapct of $\lambda$ for mAP]{
		\begin{minipage}[t]{0.25\linewidth}
			\centering
			\label{fig5e}
			\includegraphics[width=1.0\columnwidth]{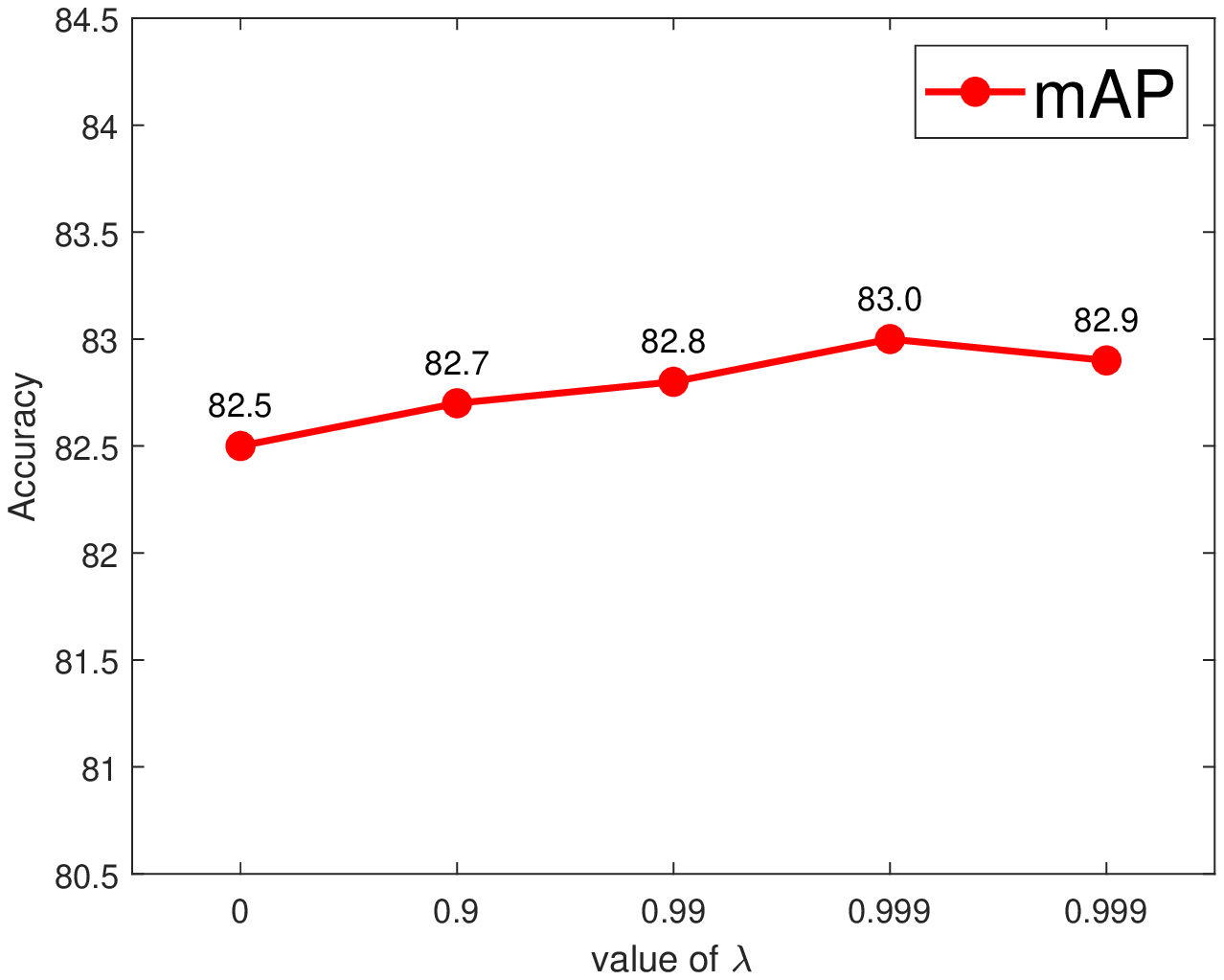}
		\end{minipage}%
	}%
	\subfloat[Imapct of $\lambda$ for Rank-1 accuracy]{
		\begin{minipage}[t]{0.25\linewidth}
			\centering
			\label{fig5f}
			\includegraphics[width=1.0\columnwidth]{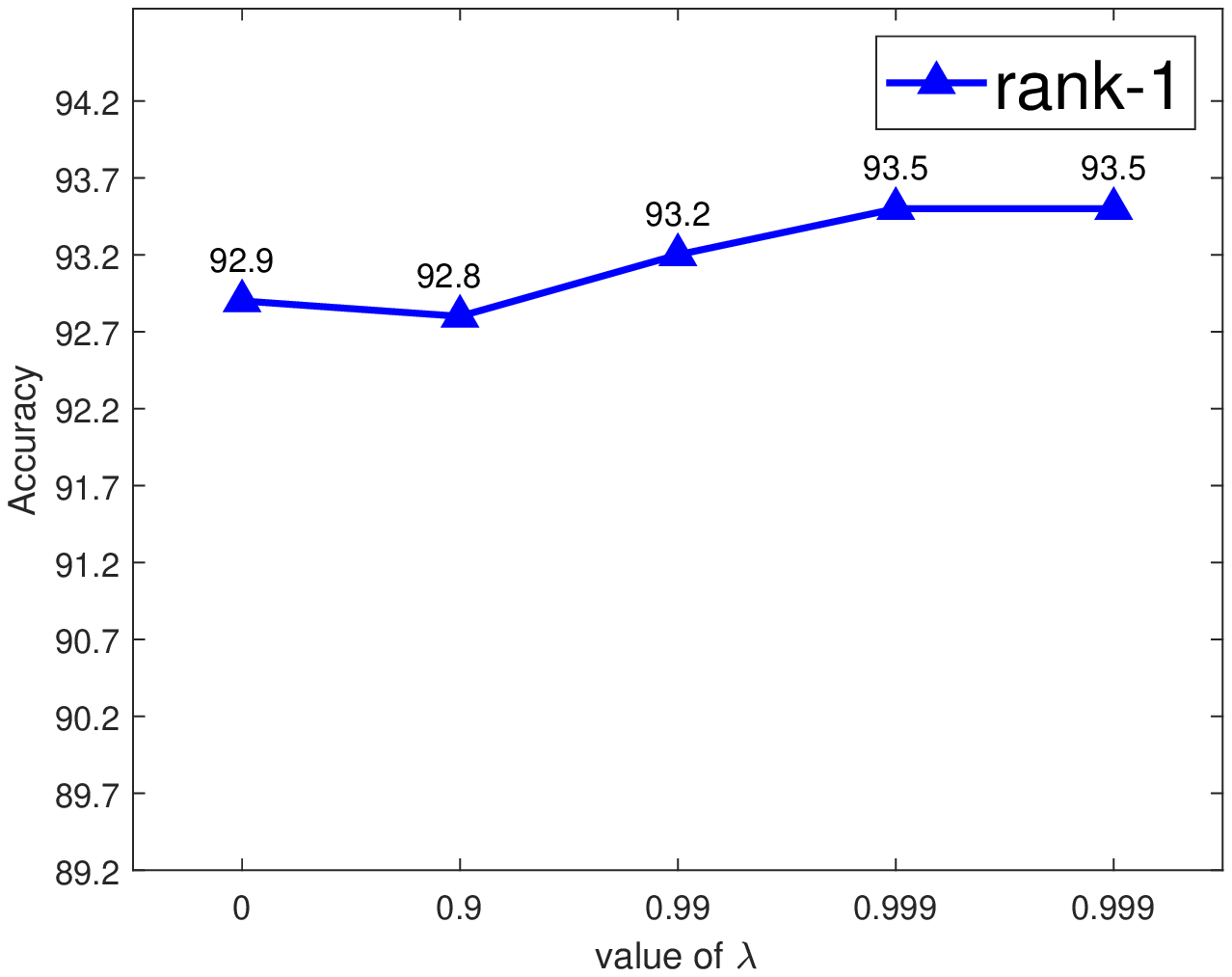}
		\end{minipage}%
	}%
	\subfloat[Imapct of $\tau$ for Rank-1 accuracy]{
		\begin{minipage}[t]{0.25\linewidth}
			\centering
			\label{fig5g}
			\includegraphics[width=1.0\columnwidth]{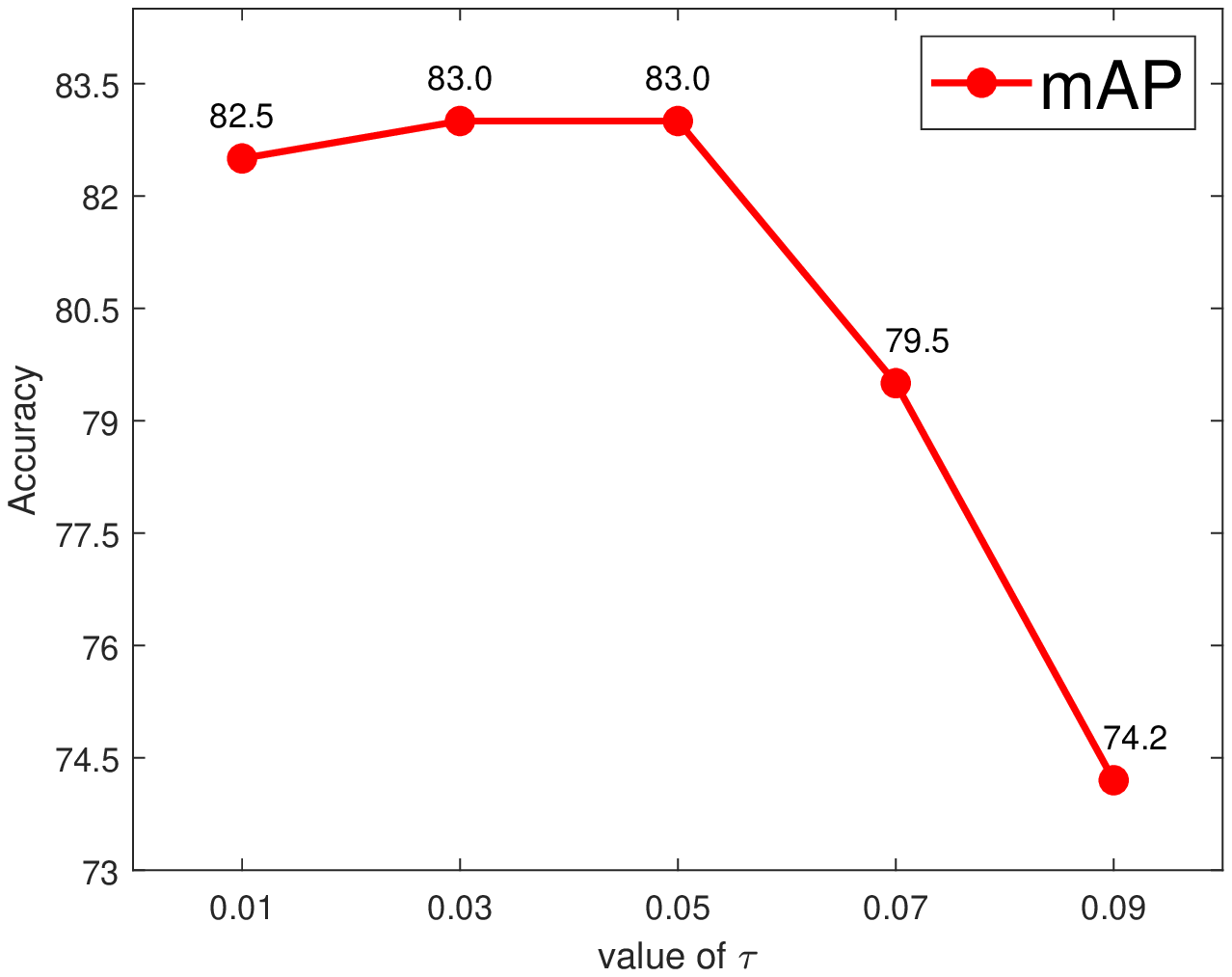}
		\end{minipage}%
	}%
	\subfloat[Imapct of $\tau$ for Rank-1 accuracy]{
		\begin{minipage}[t]{0.25\linewidth}
			\centering
			\label{fig5h}
			\includegraphics[width=1.0\columnwidth]{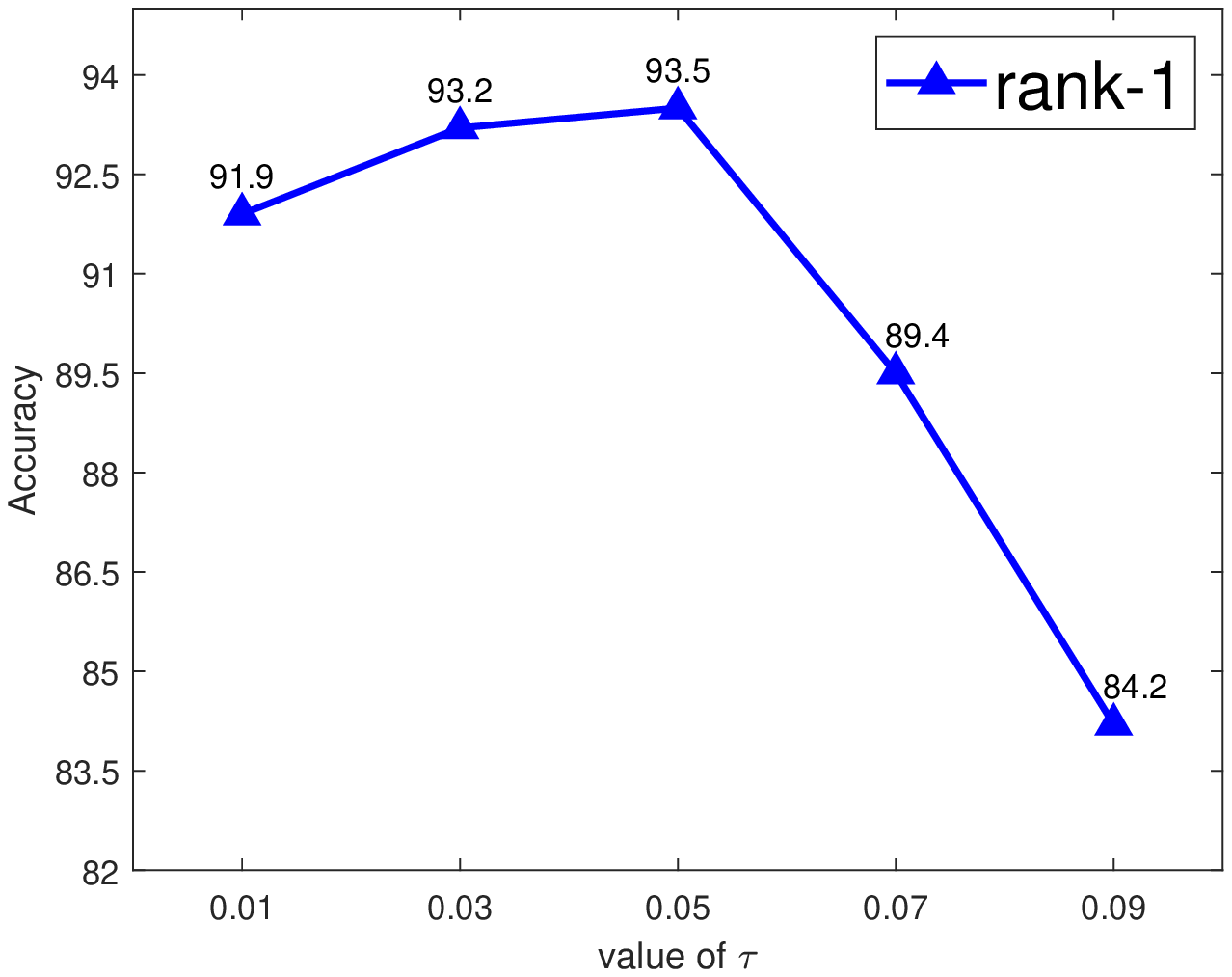}
		\end{minipage}%
	}%
	\centering
	\caption{Performance analysis of the proposed SMCR with different values of hyperparametes (\emph{i.e.,} $\alpha$, $\beta$, $\lambda$ and $\tau$) on Market-1501 $\rightarrow$ DukeMTMC-reID in terms of mAP and Rank-1 accuracies.}\label{params}
\end{figure*}

\subsection{Ablation Study}	
\subsubsection{UDA from Synthesis to Real}
To further verify the effectiveness of the constructed synthetic datasets for unsupervised domain adaptive person re-ID, we perform the UDA experiments from the labeled synthetic datasets (\emph{i.e.,} GSPR and mini-GSPR) to three unlabeled public real-world datasets: Market1501, DukeMTMC-ReID, and MSMT17. In these UDA experiments, the synthetic pretraining module is removed in the SMCR network, and we use synthetic data as the source domain to implement domain adaptation from synthetic samples to real-world datasets. The results are shown in Fig. \ref{uda_dataset}.

To be specific, the UDA performance of our SMCR network from synthetic to real-world datasets is largely improved over the current state-of-the-art results (AMBT in WACV’21) between real-world datasets, and it is comparative with the performance of our SMCR network for domain adaptive tasks between real-world datasets. Moreover, our SMCR network establishes a state-of-the-art record of the UDA task from GSPR to MSMT17 dataset. We argue that since GSPR and MSMT17 are the largest-scale person re-ID datasets in the virtual world and the real scenarios, respectively, the re-ID model can be fully trained when using GSPR instead of other small datasets. There is no apparent performance gap when leveraging mini-GSPR and GSPR as source domains, respectively, which indicates that mini-GSPR is able to help achieve remarkable performance improvements with a small data volume through delicate degign. It is worth noting the above results are gained without any manual annotation during domain adaptation, which demonstrates the application potential of our constructed synthetic datasets.

\subsubsection{Components Effectiveness}
In order to verify the effectiveness of each component in the proposed SCMR network, we create a baseline model by directly applying the backbone (\emph{IBN-ResNet50}) pretrained in the labeled source domain to conduct fine-tuning on the unlabeled target domain, while comparing the results of the baseline, the different compoenets and SCMR network on DukeMTMC-reID $\rightarrow$ Market-1501 task. All results are shown in Table \ref{ablation}.

\textbf{Necessity of Collaborative Refinement. } To explore the necessity of collaborative refinement in the proposed SMCR network, we analyze the performance of DTHR module and RIHR module in the mutually independent training manner (\emph{i.e.}, \emph{ind}) and the collaborative training way (\emph{i.e.}, \emph{col}). Significant improvement of 5.1\% and 1.3\% mAP can be obtained when we collaboratively train the devised DTHR module and RIHR module. Furthermore, it is obvious that the performance difference between DTHR module and RIHR module is significantly reduced when the independent manner is replaced with collaborative training. These results indicate the necessity of the collaborative refinement strategy for module learning and optimization.

\textbf{Impacts of Different Sub-modules. } To further analyze the value of sub-modules of DTHR module and RIHR module, firstly, we evaluate the effectiveness of cluster criteria, and considerable performance drops (35.5\% mAP) are observed when removing our cluster criteria. Then to verify the effectiveness of different backnones for the performance of our SCMR network, we replace IBN-ResNet50 with ResNet50 as the backbone of our feature encoders. There is 4.3\% mAP drop for the ResNet50 backbone. Finally, CycleGAN is considered as the domain translator to investigate the impacts of domain translator. Compared with eSPGAN, a relative decrease of 4.9\% mAP are shown for the CycleGAN domain translator. Moreover, although the above models have different sub-modules missing, all of them still significantly better than the baseline model. Overall, these results demonstrate the irreplaceability of sub-modules.

\textbf{Effectiveness of Synthetic Pretraining. } 
To verify testify the signification of our pretrained scheme for the performance of our SMCR network, firstly, we simply removed the synthetic pretraining module in our SCMR, namely SCMR (\emph {w/o syn}). Although SCMR (\emph {w/o syn}) produces a significant performance degradation compared to SCMR (\emph {GSPR}), it is still better than the current state-of-the-art results. Then SCMR (\emph {GSPR-pre}) is constructed by only leveraging the original GSPR samples as pretraining data, where the performance of SMCR (\emph {GSPR-pre}) is even lower than SMCR (\emph {w/o syn}) due to severe shift between synthetic and real-world datasets. Finally, there is only 0.6\% mAP performance difference between GSPR (\emph {GSPR}) and GSPR (\emph {mini-GSPR}), which verifies the effectiveness of mini GSPR once again.
\subsubsection{Parameter Analysis}
Finally, to evaluate the impact of different hyperparameters, we perform extensive experiments on the task of DukeMTMC-reID $\rightarrow$ Market-1501 by tuning the value of the single parameter and keeping the others fixed. All results can be found in Fig. \ref{params}.

\textbf{Balance coefficient $\alpha$. }
In Fig. \ref{fig5a} and \ref{fig5b}, we analyze the effect of the balance coefficient $\alpha$ between DTHR module and RIHR module in Eq. \ref{12}, \ref{16} and \ref{17}, where the weight for DTHR module is $\alpha$ and the weight for RIHR module is (1-$\alpha$). Our proposed SCMR network obtains the best results when $\alpha$ is set to 0.5. When $\beta$ is close to 0 or 1, the performance continues to decrease. These results show that both DTHR and RIHR modules play an indispensable role in feature learning for our SMCR network.

\textbf{Weighting factor $\beta$. }
Similar to the analysis of $\alpha$, we testify the proposed framework with different values of the weighting factor $\beta$ of collaborative losses in Eq. \ref{17}. As shown in Fig. \ref{fig5c} and \ref{fig5d}, the values of $\beta$ varies in 0, 0.001,0.01, 0.1 and 0.5. Our SMCR network gains the optimal performance with $\beta =$0.01. It can be observed that there is a significant decline when $\beta>=$ 0.01 and disastrous results were obtained with $\beta=$ 0.5. Therefore, there is no need to testify the performance of our SCMR network with $\beta$=1. These results indicate that although collaborative training has a certain positive effect on the multi-branch model, it can only be used as an auxiliary strategy.  

\begin{figure}
	\begin{center}
		\centering
		\includegraphics[width=1.0\linewidth]{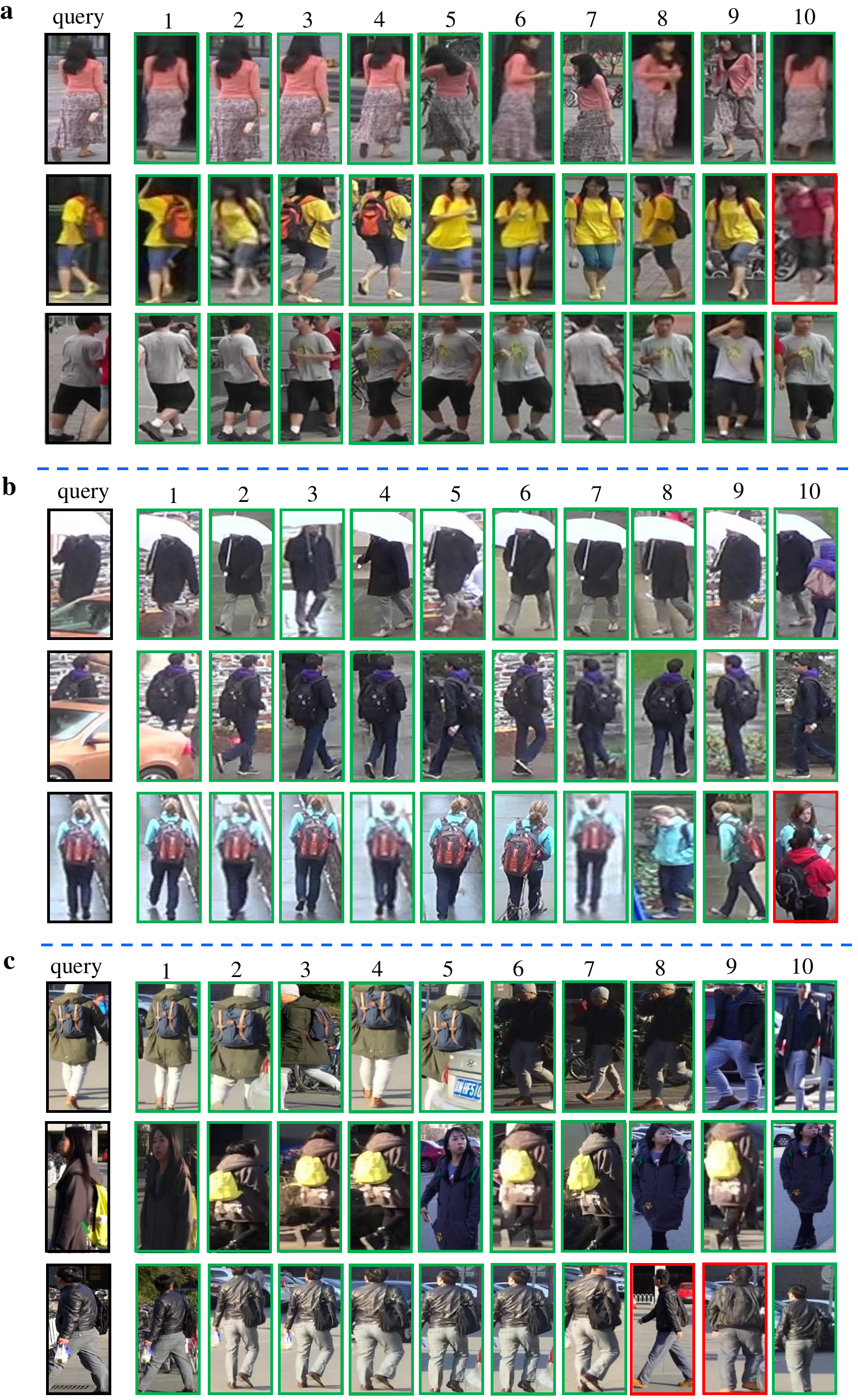}
	\end{center}
	\caption{Query examples on a$)$ Market-1501, b$)$ DukeMTMC-reID and c$)$ MSMT17. For each example, we retrieve rank-10 images from the gallery, where green boxes indicate correct retrievals and the red ones represent wrong cases.}\label{example}
\end{figure}

\textbf{Momentum Coefficient $\lambda$. }
Inspired by MoCo \cite{he2020momentum}, momentum update encoders are also introduced in our proposed method, and we use momentum coefficient $\lambda =$ 0.999 in Eq. \ref{11}. Moreover, the performance of our SMCR network is evaluated when tuning $\lambda =$ 0, 0.9, 0.99 and 0.9999, respectively. As demonstrated in Fig. \ref{fig5e} and \ref{fig5f}, we find that our framework gains the best results both on mAP and Rank-1 with $\lambda =$ 0.999, and there is a limited impact on performance for momentum update encoders.

\textbf{Temperature cofficient $\tau$. }
As shown in Fig. \ref{fig5c} and \ref{fig5d}, we further testify to the performance of our SMCR network by varying the value of temperature coefficient $\tau$ in Eq. \ref{5} and \ref{9}. Similar to recent methods \cite{zhong2019invariance}, \cite{zhong2020learning}, using temperature function, our method obtains the optimal results with $\tau =$ 0.05, and there is a dramatic decline when $\tau >=$ 0.05.

\subsubsection{Qualitative Analysis}
Finally, to qualitatively analyze the performance of the proposed SMCR network, we show some query examples of SMCR network on Market-1501 \cite{zheng2015scalable}, DukeMTMC-reID \cite{ristani2016performance} and MSMT17 \cite{wei2018person}. As shown in Fig. \ref{example}, the SMCR network achieves remarkable retrieval performance even for the challenging person samples, where the proposed method can correctly retrieve the person images with significant pose and viewpoint variations, and it is robust to occlusions and background clutter.

\section{Conclusion}\label{sec:conclusion}
In this work, we focus on alleviating labeled data severe dependence and poor performance of domain adaptive tasks in the field of person re-identification via synthetic data. To this end, we first develop an automatic data collector/labeler and construct two synthetic person re-ID datasets with different scales, which do not need any manual labels and free humans from heavy data collections and annotations. In order to further exploit the generated synthetic data, we propose a synthesis-based multi-domain collaborative refinement (SMCR) network to sufficiently absorb the valuable knowledge from multiple domains by performing synthetic pretraining and collaborative refinement. Extensive experiments demonstrate the proposed method achieves consistent performance gains over the state-of-the-art approaches on multiple unsupervised domain adaptation tasks of person re-ID. 

In future work, we will focus on self-supervised learning for the re-ID task, which often needs more data and may implement better results. Moreover, it is desirable to generalize domain adaptive learning via synthetic data from image-based to video-based.


\bibliographystyle{IEEEtran}
\bibliography{tip}

\begin{IEEEbiography}[{\includegraphics[width=1in,height=1.25in,clip,keepaspectratio]{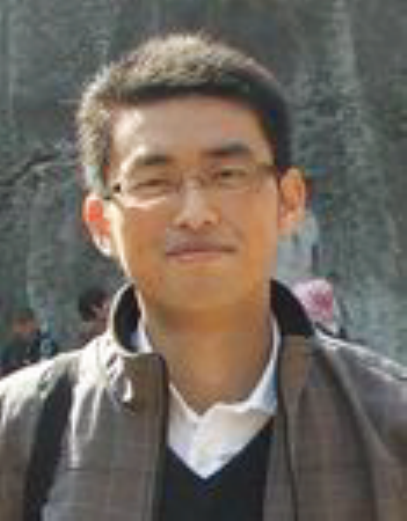}}]{Qi Wang}
	(M’15-SM’15) received the B.E. degree in automation and the Ph.D. degree in
	pattern recognition and intelligent systems from
	the University of Science and Technology of
	China, Hefei, China, in 2005 and 2010, respectively. He is currently a Professor with the School of Artificial Intelligence, OPtics and Electronics (iOPEN), Northwestern Polytechnical University, Xi'an 710072, P.R. China. His research interests include computer vision and pattern recognition.
\end{IEEEbiography}

\begin{IEEEbiography}[{\includegraphics[width=1in,height=1.25in,clip,keepaspectratio]{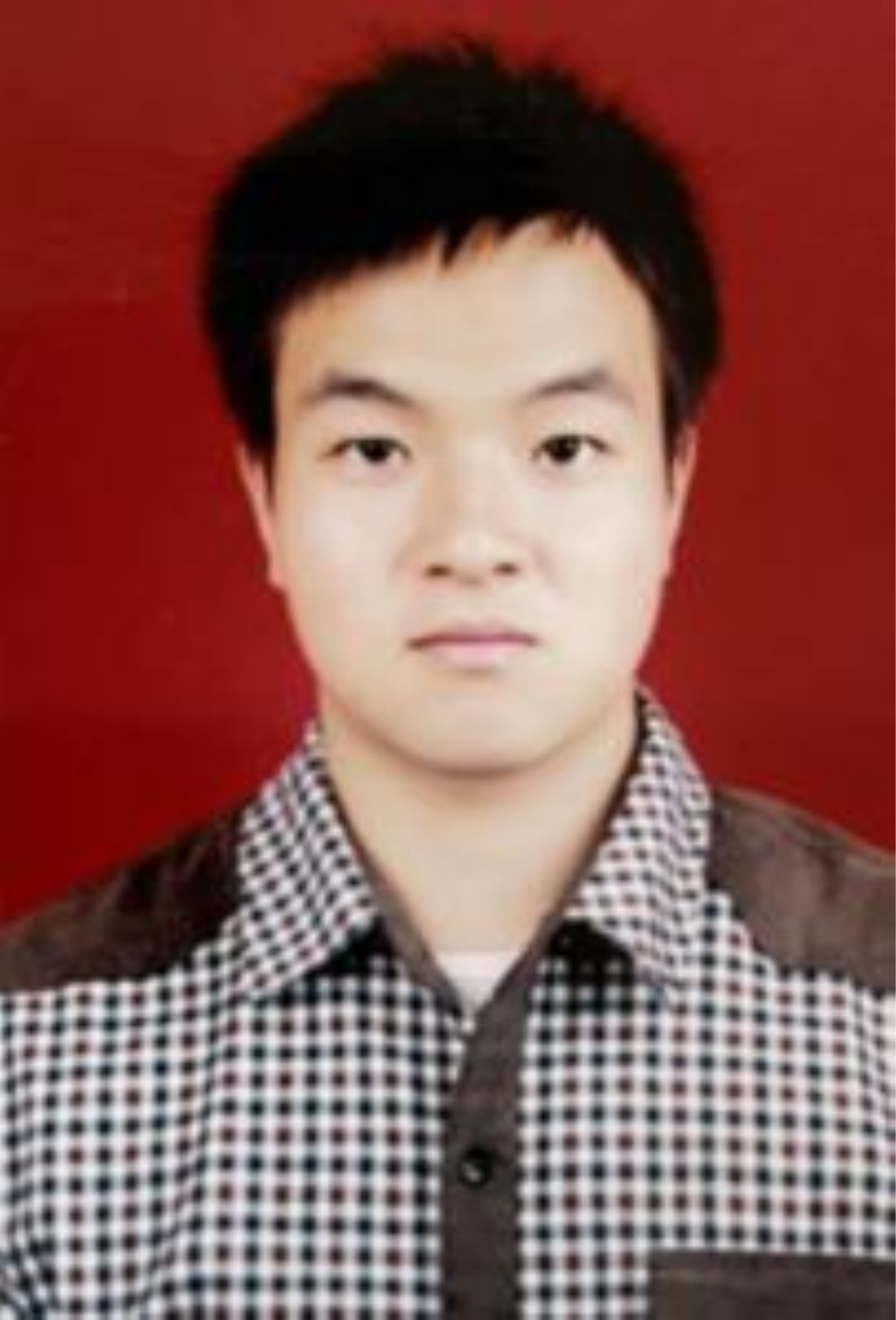}}]{Sikai Bai}
	received the B.E. degree in software engineering from China University Of Geosciences, Wuhan, China, in 2019. He is currently working toward the M.S. degree in the School of Computer Science and School of Artificial Intelligence, OPtics and Electronics (iOPEN), Northwestern Polytechnical University, Xi'an 710072, P.R. China. His research interests include computer vision and pattern recognition.
\end{IEEEbiography}

\begin{IEEEbiography}[{\includegraphics[width=1in,height=1.25in,clip,keepaspectratio]{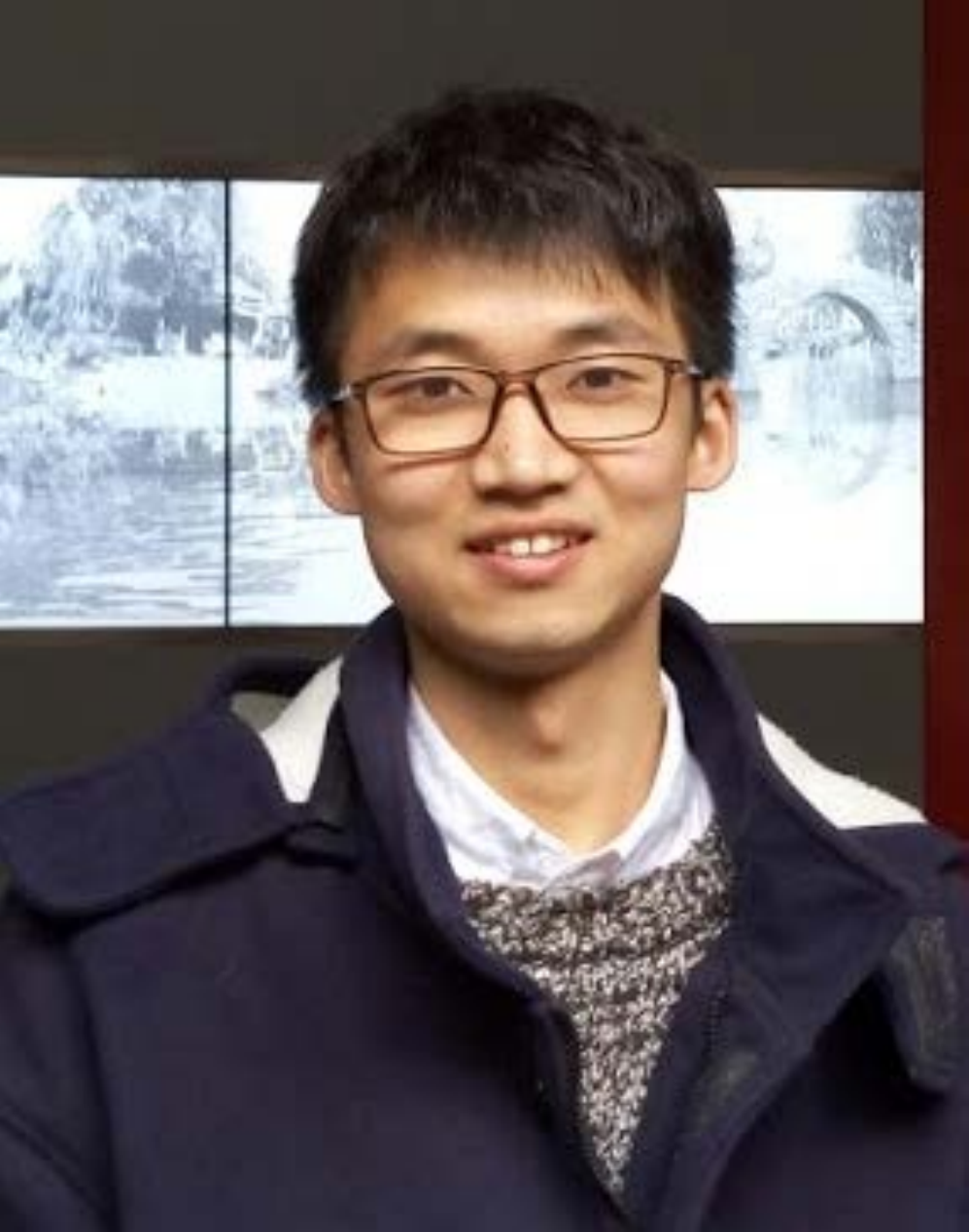}}]{Junyu Gao}
	received the B.E. degree and the Ph.D. degree in computer
	science and technology from the Northwestern Polytechnical University, Xi’an 710072, Shaanxi, P. R.
	China, in 2015 and 2020, respectively. He is currently an Assistant Professor with the School of Artificial Intelligence, OPtics and Electronics (iOPEN), Northwestern Polytechnical University, Xi'an 710072, P.R. China. His research interests include
	computer vision and pattern recognition.
\end{IEEEbiography}

\begin{IEEEbiography}[{\includegraphics[width=1in,height=1.25in,clip,keepaspectratio]{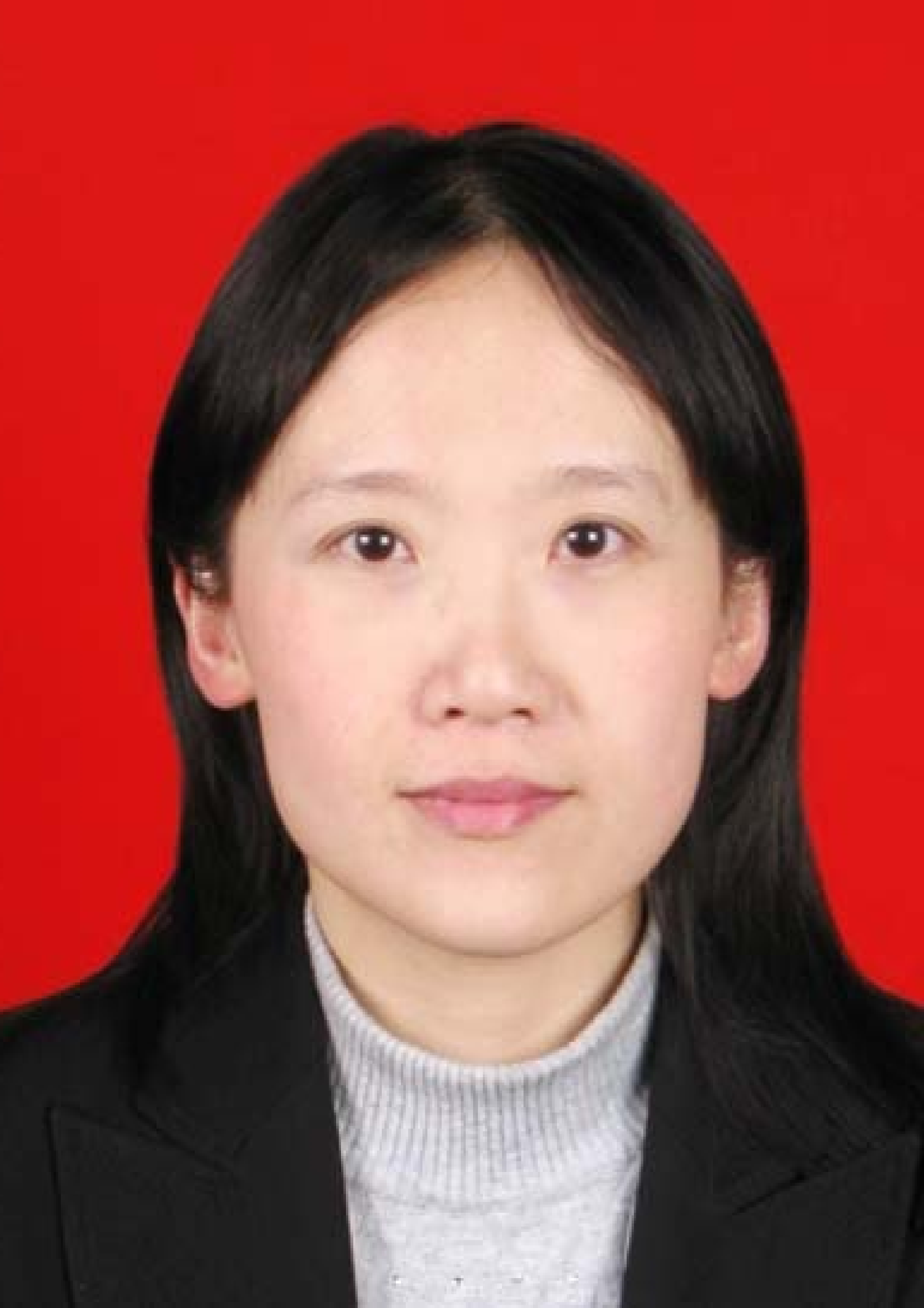}}]{Yuan Yuan}
	(M’05-SM’09) is currently a Full Professor with the School of Artificial Intelligence, OPtics and Electronics (iOPEN),
	Northwestern Polytechnical University, Xi’an 710072, Shaanxi, P. R.
	China. She has authored or co-authored over 150 papers,
	including about 100 in reputable journals, such as the
	IEEE TRANSACTIONS AND PATTERN RECOGNITION, as well as the conference papers in CVPR,
	BMVC, ICIP, and ICASSP. Her current research
	interests include visual information processing and
	image/video content analysis.
\end{IEEEbiography}

\vfill

\end{document}